%% file: sn-article.tex
\documentclass[sn-mathphys, iicol]{sn-jnl}

\usepackage{graphicx}%
\usepackage{amsmath,amssymb,amsfonts}%
\usepackage{amsthm}%
\usepackage{mathrsfs}%
\usepackage[title]{appendix}%
\usepackage{xcolor}%
\usepackage{textcomp}%
\usepackage{manyfoot}%
\usepackage{booktabs}%
\usepackage{algorithm}%
\usepackage{algorithmicx}%
\usepackage{algpseudocode}%
\usepackage{listings}%
\usepackage[normalem]{ulem}
\usepackage{relsize}
\usepackage{array,multirow}
\usepackage{tabulary}
\usepackage{booktabs}
\usepackage{caption} 
\usepackage{pifont} 
\usepackage{comment} 
\newcolumntype{C}[1]{>{\centering\arraybackslash}p{#1}}

\definecolor{coolgreen}{RGB}{6,169,77}
\definecolor{coolred}{RGB}{244,49,49}

\usepackage{caption}
\usepackage{arydshln}
\usepackage{adjustbox}
\usepackage{makecell}

\usepackage{orcidlink}

\usepackage{ifthen}
\usepackage{xstring}
\usepackage{etoolbox}
\usepackage{pgf}

\definecolor{mygreen}{HTML}{77dd77}  
\definecolor{myred}{HTML}{f3443f} 
\definecolor{myyellow}{HTML}{ffff66}

\newcommand{\gradient}[3]{%
    \IfStrEq{#1}{office31_resnet50}{\def \MinNumber{57} \def\MidNumber{85} \def\MaxNumber{89.4}}{}\par
    \IfStrEq{#1}{office31_vit}{\def\MinNumber{52.8} \def\MidNumber{92} \def\MaxNumber{94.9}}{}\par

    \IfStrEq{#1}{officehome_resnet50}{\def \MinNumber{22} \def\MidNumber{70} \def\MaxNumber{73}}{}\par
    \IfStrEq{#1}{officehome_vit}{\def\MinNumber{44} \def\MidNumber{82} \def\MaxNumber{88}}{}\par

    \IfStrEq{#1}{adaptiope_resnet50}{\def \MinNumber{4.9} \def\MidNumber{55} \def\MaxNumber{72}}{}\par
    \IfStrEq{#1}{adaptiope_vit}{\def\MinNumber{23} \def\MidNumber{84} \def\MaxNumber{90}}{}\par

    \IfStrEq{#1}{imageclef_resnet50}{\def \MinNumber{67.2} \def\MidNumber{82} \def\MaxNumber{83.4}}{}\par
    \IfStrEq{#1}{imageclef_vit}{\def\MinNumber{73.5} \def\MidNumber{87} \def\MaxNumber{88.5}}{}\par

    \ifdim #2 pt > \MidNumber pt 
        \pgfmathsetmacro{\PercentColor}{int(round((100.0*(#2 - \MidNumber)/(\MaxNumber-\MidNumber))))} 
        \colorbox{mygreen!\PercentColor!myyellow}{\makebox(22,12){\makecell{#2 \\  $\scriptstyle \boldsymbol{\pm}$ \footnotesize #3}}}
    \else 
        \pgfmathsetmacro{\PercentColor}{int(round((100.0*(\MidNumber - #2)/(\MidNumber-\MinNumber))))} 
        \colorbox{myred!\PercentColor!myyellow}{\makebox(22,12){\makecell{#2 \\  $\scriptstyle \boldsymbol{\pm}$ \footnotesize #3}}}
    \fi  
}%

\newcommand{\gradientfull}[3]{%
    \IfStrEq{#1}{office31_resnet50}{\def \MinNumber{57} \def\MidNumber{85} \def\MaxNumber{89.4}}{}\par
    \IfStrEq{#1}{office31_vit}{\def\MinNumber{52.8} \def\MidNumber{92} \def\MaxNumber{94.9}}{}\par

    \IfStrEq{#1}{officehome_resnet50}{\def \MinNumber{22} \def\MidNumber{70} \def\MaxNumber{73}}{}\par
    \IfStrEq{#1}{officehome_vit}{\def\MinNumber{44} \def\MidNumber{82} \def\MaxNumber{88}}{}\par

    \IfStrEq{#1}{adaptiope_resnet50}{\def \MinNumber{4.9} \def\MidNumber{55} \def\MaxNumber{72}}{}\par
    \IfStrEq{#1}{adaptiope_vit}{\def\MinNumber{23} \def\MidNumber{84} \def\MaxNumber{90}}{}\par

    \IfStrEq{#1}{imageclef_resnet50}{\def \MinNumber{67.2} \def\MidNumber{82} \def\MaxNumber{83.4}}{}\par
    \IfStrEq{#1}{imageclef_vit}{\def\MinNumber{73.5} \def\MidNumber{87} \def\MaxNumber{88.5}}{}\par

    \ifdim #2 pt > \MidNumber pt 
        \pgfmathsetmacro{\PercentColor}{int(round((100.0*(#2 - \MidNumber)/(\MaxNumber-\MidNumber))))} 
        \colorbox{mygreen!\PercentColor!myyellow}{\makebox(22,106){\makecell{#2 \\  $\scriptscriptstyle \boldsymbol{\pm}$ \footnotesize #3}}}
    \else 
        \pgfmathsetmacro{\PercentColor}{int(round((100.0*(\MidNumber - #2)/(\MidNumber-\MinNumber))))} 
        \colorbox{myred!\PercentColor!myyellow}{\makebox(22,106){\makecell{#2 \\  $\scriptscriptstyle \boldsymbol{\pm}$ \footnotesize #3}}}
    \fi  
}%

\newcommand{\gradientsingle}[2]{%

    \IfStrEq{#1}{aad_hparams}{\def \MinNumber{69.5} \def\MidNumber{75.3} \def\MaxNumber{76.35}}{}\par
    \IfStrEq{#1}{nrc_hparams}{\def\MinNumber{74.3} \def\MidNumber{76.5} \def\MaxNumber{77.5}}{}\par
    
    \ifdim #2 pt > \MidNumber pt 
        \pgfmathsetmacro{\PercentColor}{int(round((100.0*(#2 - \MidNumber)/(\MaxNumber-\MidNumber))))} 
        \colorbox{mygreen!\PercentColor!myyellow}{\makebox(30,5){#2}}
    \else 
        \pgfmathsetmacro{\PercentColor}{int(round((100.0*(\MidNumber - #2)/(\MidNumber-\MinNumber))))} 
        \colorbox{myred!\PercentColor!myyellow}{\makebox(30,5){#2}}
    \fi  
}%

\newcommand{\n}{ \\[0.4em] }

\newcommand{\newtabsection}{ \hspace{0.6em}&\hspace{-0.1em}}

\theoremstyle{thmstyletwo}%

\theoremstyle{thmstylethree}%

\begin{document}

\title[Article Title]{Key~Design~Choices~in~Source-Free Unsupervised~Domain~Adaptation:
An~In-depth~Empirical~Analysis}



\author*[1,2,3]{\fnm{Andrea} \sur{Maracani} \orcidlink{0000-0002-6217-8731}}\email{andreamaracani@gmail.com}

\author[4]{\fnm{Raffaello} \sur{Camoriano} \orcidlink{0000-0002-8890-2732}}\email{raffaello.camoriano@polito.it}
\equalcont{These authors contributed equally to this work.}

\author[1]{\fnm{Elisa} \sur{Maiettini} \orcidlink{0000-0002-0127-3014}}\email{elisa.maiettini@iit.it}
\equalcont{These authors contributed equally to this work.}

\author[6, 7]{\fnm{Davide} \sur{Talon} \orcidlink{0009-0003-6029-1532}}\email{talon.davide@gmail.com}

\author[2,3,5]{\fnm{Lorenzo} \sur{Rosasco} \orcidlink{0000-0003-3098-383X}}\email{lrosasco@mit.edu}

\author[1]{\fnm{Lorenzo} \sur{Natale} \orcidlink{0000-0002-8777-5233}}\email{lorenzo.natale@iit.it}

\affil[1]{\orgdiv{HSP}, \orgname{Istituto Italiano di Tecnologia}, \orgaddress{\city{Genoa}, \country{Italy}}}

\affil[2]{\orgdiv{DIBRIS}, \orgname{Università di Genova}, \orgaddress{\city{Genoa}, \country{Italy}}}

\affil[3]{\orgdiv{MaLGa Center}, \orgname{Università di Genova}, \orgaddress{\city{Genoa}, \country{Italy}}}

\affil[4]{\orgdiv{VANDAL Laboratory}, \orgname{Politecnico di Torino}, \orgaddress{\city{Turin}, \country{Italy}}}

\affil[5]{\orgdiv{IIT@MIT}, \orgname{Istituto Italiano di Tecnologia}, \orgaddress{ \city{Genoa}, \country{Italy}}}

\affil[6]{\orgdiv{PAVIS}, \orgname{Istituto Italiano di Tecnologia}, \orgaddress{\city{Genoa}, \country{Italy}}}
\affil[7]{\orgdiv{DITEN}, \orgname{Università di Genova}, \orgaddress{\city{Genoa}, \country{Italy}}}

\abstract{

This study provides a comprehensive benchmark framework for Source-Free Unsupervised Domain Adaptation (SF-UDA) in image classification, aiming to achieve a rigorous empirical understanding of the complex relationships between multiple key design factors in SF-UDA methods.
The study empirically examines a diverse set of SF-UDA  techniques, assessing their consistency across datasets, sensitivity to specific hyperparameters, and applicability across different families of backbone architectures.
Moreover, it exhaustively evaluates pre-training datasets and strategies, particularly focusing on both supervised and self-supervised methods, as well as the impact of fine-tuning on the source domain.
Our analysis also highlights gaps in existing benchmark practices, guiding SF-UDA research towards more effective and general approaches. 
It emphasizes the importance of backbone architecture and pre-training dataset selection on SF-UDA performance, serving as an essential reference and providing key insights.
Lastly, we release the source code of our experimental framework.
This facilitates the construction, training, and testing of SF-UDA methods, enabling systematic large-scale experimental analysis and supporting further research efforts in this field.

}


\maketitle

\section{Introduction}
\label{sec:introduction}
\input{Introduction}

\section{Related Work}
\label{sec:related_work}
\input{Related_work}
\section{Methods}
\label{sec:methods_datasets_intro}
\input{Methods}
\section{Benchmarking Framework}
\label{sec:framework}
\input{Framework}

\section{Analysis of SF-UDA methods}
\label{sec:methods}
\input{experiments_methods}

\section{Impact of pre-training and backbone selection}
\label{sec:pretraining}
\input{experiments_pretraining}

\section{Analysis of the impact of fine-tuning}
\label{sec:finetuning}
\input{experiments_finetuning}

\section{Discussion}
\label{sec:discussions}
\input{Discussions}

\section{Conclusion}
\label{sec:conclusion}
\input{Conclusions} 

\section*{Data availability}
\input{data_availability}

\section*{Acknowledgements}
\input{acknowledgements}

\bibliography{bibliography}

\end{document}

%% file: Introduction.tex
Deep Learning has established itself as the leading approach to tackle most computer vision tasks.  
However, the performance of Deep Neural Networks (DNNs) is largely dependent on the availability of large-scale annotated datasets, which may be costly to acquire and also challenging for specialized tasks.
To address this issue, the prevalent strategy is to initially pre-train model weights on extensive datasets and then \textit{fine-tune} them for specific tasks using a smaller set of labeled examples~\citep{huh2016makes, yosinski2014transferable, chu2016best}.
This approach adheres to the transfer learning~\citep{Zhuang2021Comprehensive} paradigm: expertise acquired on one task can subsequently enhance learning performance in related, yet distinct downstream tasks.

\noindent Additionally, DNNs exhibit strong performance only when the training and test datasets are drawn from the same distribution.
However, in practical applications several factors, such as environmental variations or dataset biases, lead to a domain shift between training (\textit{source domain}) and test data (\textit{target domain}), potentially causing a significant performance degradation.  
To bridge the gap between source and target domains, \textbf{Unsupervised Domain Adaptation} (UDA) jointly employs labeled data from the source domain and unlabeled data from the target domain to improve model adaptation~\citep{wilson2020survey}.
In particular, in this work we focus on the more challenging \textbf{Source-Free Unsupervised Domain Adaptation} (SF-UDA) setting~\citep{liang2020we}. SF-UDA presents more constraints than UDA, involving a two-stage training, here referred as \textit{double transfer}: (i)~a pre-trained model is first fine-tuned on a labeled source domain, and (ii)~it is subsequently adapted by employing only the unlabeled data from the target domain with no further access to any source data.
This setting is especially useful in applications where access to source data is constrained due to privacy, communication, and storage issues.

\noindent \cite{kim2022broad} conducted a rigorous study on strategies commonly adopted in UDA. In particular design choices like architecture and pre-training strategy have been explored in the context of UDA. Their findings indicate that some recent methodologies may be tailored to excel on specific benchmarks: when certain modifications are applied (e.g., the architecture is changed), such methods under-perform compared to earlier approaches. 
Motivated by these analyses for UDA, our research centers on the SF-UDA setting, with a focus on understanding the contribution of each phase in its characteristic \textit{double transfer}.
Our analysis extends beyond standard benchmark assessments, providing an in-depth characterization of multiple design components, ranging from pre-training methods and adaptation strategies to hyper-parameter sensitivity, and evaluating the effectiveness of fine-tuning on the source domain.
Through our analysis, we provide a novel perspective into the strengths and limitations of SF-UDA pipelines, setting a foundation for future progress in the field.
The main contributions of this work are as follows:

\begin{itemize}

\item We propose a benchmark framework to evaluate SF-UDA methods focusing on their general applicability in different experimental settings, their reproducibility, robustness, and failure rates. 

\item We analyze the influence of the backbone and pre-training dataset choices on the final SF-UDA performance.
Our results show a high correlation between the ImageNet top-1 accuracy, Out-of-Distribution Generalization (ODG), and SF-UDA performance across hundreds of architectures, including Convolutional Neural Networks and Vision Transformers.
Furthermore, we analyze the performance of Self-Supervised pre-training strategies, which are gaining momentum in the current literature.

\item While fine-tuning the backbone  on the source domain is common practice in SF-UDA, we show that in some scenarios this  leads to severe performance degradation. 
Additionally, we highlight the marked effect of normalization layer selection on SF-UDA perfomance. 
Architectures with Layer Normalization consistently outpace those employing Batch Normalization (Sec.~\ref{sec:finetuning}).
\end{itemize}

The following sections discuss related work (Sec.~\ref{sec:related_work}), detail our benchmark approach (Sec. \ref{sec:methods_datasets_intro} and \ref{sec:framework}), present experimental results (Sec.~\ref{sec:methods}, \ref{sec:pretraining} and \ref{sec:finetuning}), and conclude with insights on findings and directions for future research (Sec.~\ref{sec:discussions} and~\ref{sec:conclusion}).

%% file: Related_work.tex
\textbf{Unsupervised Domain Adaptation (UDA).} 
DNNs and other Machine Learning models typically operate under the assumption that their training and test datasets are drawn from the same distribution~\citep{wilson2020survey}. This assumption, however, often does not hold in practical applications, leading to significant performance drops. 
\noindent To address these challenges, Unsupervised Domain Adaptation (UDA) has been proposed. 
UDA leverages a labeled source domain alongside an unlabeled target domain, aiming to optimize model performance on the latter.
Early theoretical works~\citep{ben2006analysis, ben2010theory, mansour2009domain} formed the foundation for a variety of UDA algorithms, applicable across fields such as time series data analysis~\citep{liu2021adversarial}, Natural Language Processing~\citep{ramponi2020neural}, Sentiment Analysis~\citep{dai2020adversarial}, and computer vision tasks such as video analysis~\citep{xu2022video}, image classification~\citep{madadi2020deep, li2021unsupervised}, object detection~\citep{oza2023unsupervised, xu2016hierarchical}, and semantic segmentation~\citep{toldo2020unsupervised}.

\noindent Specifically, UDA has been extensively studied in image classification. 
Proposed techniques include Adversarial Training~\citep{ganin2016domain}, Maximum Mean Discrepancy Minimization~\citep{kang2019contrastive}, Bi-directional Matching~\citep{na2021fixbi}, Margin Disparity Discrepancy~\citep{zhang2019bridging} and Contrastive Learning~\citep{wang2022cross}. 
These methods proved successful at enabling DNNs to bridge the  gap between source and target domains, 
thereby enhancing their adaptability and accuracy in diverse and changing environments.

\noindent \textbf{Source-Free Unsupervised Domain Adaptation (SF-UDA).} 
Our study specifically focuses on SF-UDA, a more constrained subset of UDA where there is no granted access to source data during adaptation.
SF-UDA emerges as a suitable setting in scenarios where, due to concerns such as intellectual property, privacy, or memory limitations, the labeled source domain is not available during the adaptation phase. 
The foundation of SF-UDA is Hypothesis Transfer Learning~\citep{kuzborskij2013stability}, which inspired the first SF-UDA methods to adapt DNNs in image classification tasks, such as SHOT~\citep{liang2020we}.
These early methods showed competitive performance compared to  UDA approaches.
Further works introduced a variety of SF-UDA algorithms.
These include generative modeling~\citep{li2020model}, techniques focusing on entropy minimization, transferring Batch Normalization statistics~\citep{hou2020source}, creating surrogate source domains during adaptation~\citep{ding2023proxymix}, employing self-distillation techniques~\citep{liu2021graph}, and leveraging the latent structure of source-trained models for adaptation~\citep{yang2022attracting, yang2021exploiting}. Contrastive learning has also been employed in this context~\citep{agarwal2022unsupervised}. A comprehensive review of contemporary SF-UDA methods is available in \cite{fang2022source}, and the specific methods explored in our study are detailed in Section \ref{subsec:sfuda_methods}.

\noindent \textbf{Experimental Studies.} The challenges surrounding DNNs training, given their computational, time, and data-intensive requirements, have prompted significant research efforts.
Their main goals have been to extract representations suitable for effective transfer to new tasks and to identify the key components enabling more efficient architectures and training methods.
Past efforts have explored many aspects of transfer learning, as shown in works by  \cite{chu2016best}, \cite{huh2016makes},  and~\cite{kornblith2019better}.
In the  UDA setting, \cite{zhang2020impact} explore model selection based on ImageNet performance, while  \cite{kim2022broad} have analyze various pre-training techniques for Domain Generalization and UDA.\\
Conversely, this work focuses on the potentials and limitations of SF-UDA, a setting gaining traction for its ability to rival traditional UDA in performance while enforcing stricter data constraints.
Our comprehensive analysis decouples and studies in depth the properties of each phase in the double transfer process: transitioning from pre-training to the source domain and, then, from the source to the target domain. 
This level of \textit{modular} decomposition of the SF-UFA pipeline is inherently unfeasible in traditional UDA where the adaptation is performed using the source and target domains, concurrently, in a single phase.
Finally, along with our results, we release the experimental framework code, which is a valuable to facilitate future research. 
In particular, the code not only enables the replication of our results, but it also serves as a foundation for further empirical investigation, as it can be effortlessly expanded with novel methods, architectures, and datasets.

%% file: Methods.tex
This section describes the high-level SF-UDA pipeline and introduces the SF-UDA approaches included in our study.

\subsection{SF-UDA pipeline}\label{subsec:sfuda_pipeline}

In SF-UDA, two data distributions (domains) are defined: the \textbf{source domain} and the \textbf{target domain}. Under the Closed-set assumption, images from different domains may differ in style, yet share the same label set. 
Model training is divided into different phases based on the data availability. 
In particular, the full process can be summarized in \textbf{three main stages}:
\begin{enumerate}
    \item A model (a DNN backbone) is initially \textbf{pre-trained} on a large dataset, e.g., ImageNet or Imagenet21K \citep{deng2009imagenet};
    \item \textbf{First transfer}: the model weights can potentially be fine-tuned (FT) using the labeled source dataset;
    \item \textbf{Second transfer}: unlabeled images from the target domain are employed for model adaptation to the new domain. Note that no access to the source data is available at this stage.
\end{enumerate}

\begin{figure*}[t]
    \centering
    \includegraphics[width=0.80\linewidth]{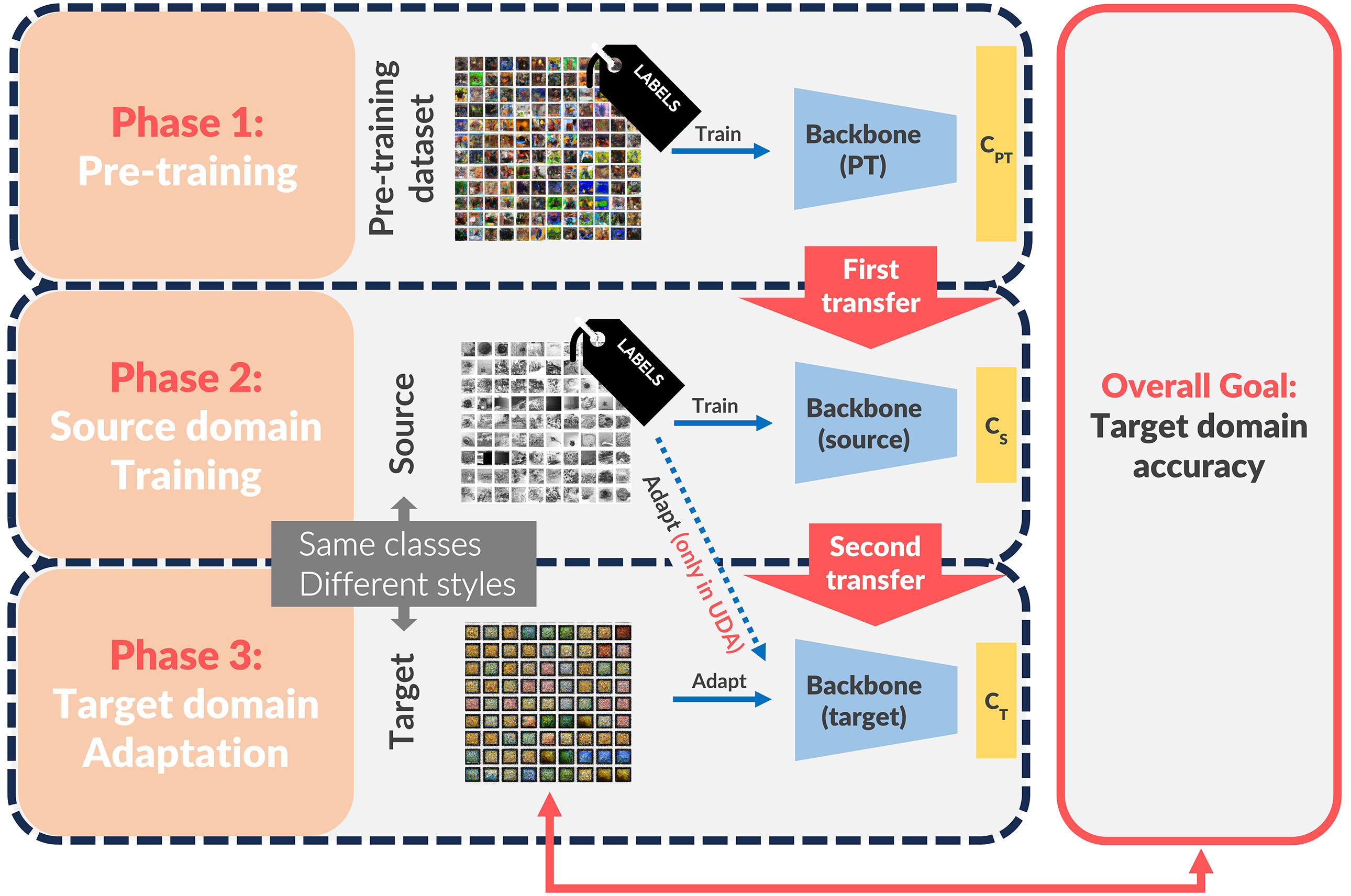}
    \caption{\textbf{SF-UDA pipeline.} In this work, we meticulously analyze the SF-UDA pipeline. The process begins \textbf{pre-training} a backbone (along with its classifier, $C_{PT}$) on a large dataset, e.g., ImageNet (see Sec. \ref{sec:pretraining}). This is followed by the \textbf{first transfer} phase, where labeled source data is used to refine the backbone and, possibly, train a classifier for the new task (see Sec. \ref{sec:finetuning}). Then, the \textbf{second transfer} phase happens, leveraging unlabeled target data to adapt the model in the target domain (see Sec. \ref{sec:methods}).
}
    \label{fig:pipeline}
\end{figure*}

\noindent The outlined modular pipeline is illustrated in Fig.~\ref{fig:pipeline}. This structure facilitates a methodical evaluation of its individual components: the strategy for unsupervised adaptation using target images (Sec.~\ref{sec:methods}), the criteria for selecting the pre-training dataset and methodology (Sec.~\ref{sec:pretraining}), and the effectiveness and issues related to fine-tuning the model's weights on the source domain (Sec.~\ref{sec:finetuning}).
This work includes a sequence of controlled experiments designed to isolate and evaluate the impact of each individual component on the final outcomes.

\smallskip

\subsection{SF-UDA methods}
\label{subsec:sfuda_methods}

To quantitatively assess the performance of SF-UDA techniques across a wide range of configurations, we conduct a detailed software re-implementation guided by the original research papers and their associated code.
Importantly, we add the options to employ many different backbones and perform distributed training.
In particular, for our analyses we select a set of representative SF-UDA methods: SCA (Sec.~\ref{sub:sca}), SHOT~\citep{liang2020we} (Sec.~\ref{sub:shot}),  NRC~\citep{yang2021exploiting} (Sec.~\ref{sub:nrc}), AAD~\citep{yang2022attracting} (Sec.~\ref{sub:aad}) and PCSR~\citep{guan2022polycentric} (Sec.~\ref{sub:pcsr}).
Several other SF-UDA approaches exist in the literature~\citep{fang2022source}. Still, we focus on the aforementioned methods for the following reasons:

\begin{itemize}
\item Despite our meticulous implementation process, it was not possible to replicate the results of some methods as presented in their original papers, indicating a sensitivity to particular experimental configurations.
\item Some methods were excluded because their official source code was unavailable (e.g., \cite{xia2021adaptive, wang2022exploring}) or their corresponding papers lacked comprehensive training or implementation details.
\item We omitted certain methods designed for specific architectures (e.g., \cite{hou2020source, liu2023source}) or those requiring unique architectural modifications or additions (e.g., \cite{yang2021transformer, li2020model}) or external data/resources (e.g., methods designed for multi-source SF-UDA as \cite{dong2021confident}). These methods were not pertinent to our wide-ranging architectural evaluation.
\end{itemize}

\noindent In the next paragraphs, we briefly describe each SF-UDA method used in our study.

\subsubsection{Simple Class Alignment}
\label{sub:sca}
\noindent Simple Class Alignment (\textbf{SCA}) is a simple yet effective method, often employed as a part of other domain adaptation algorithms, such as CAN~\citep{kang2019contrastive} and SHOT~\citep{liang2020we}. However, its use as a standalone method has not been explored in the literature, and its specific role within multistage algorithms still requires close examination.
Furthermore, despite its simplicity, we observe that in some settings it reaches state-of-the-art results (see Sec.~\ref{sub:remarks}). For these reasons, we included it in our study.
The main algorithmic steps of SCA are:

\begin{enumerate}
    \item \textbf{Prototype Creation}: generate a representative vector (prototype) for each class in the source domain (for instance by averaging the feature vectors of samples in that class).
    \item \textbf{K-Means Initialization}: use the source prototypes as starting centroids for K-Means clustering on the unlabeled target domain data.
    \item \textbf{Prototype Adaptation and Classification}: apply spherical K-Means~\citep{hornik2012spherical} to fit prototypes to target domain feature vectors.
    The final prototypes are then employed by a 1-Nearest Neighbor classifier: each target sample is assigned the class of the closest prototype, employing the cosine similarity as metric.
\end{enumerate}

\noindent SCA's distinguishing feature is its capability to provide a classifier for the target domain by aligning source prototypes with the target latent vectors distribution. 
This enables its use with feature representations from a pre-trained model with no additional refinement.
Hence, SCA emerges as a resource-efficient SF-UDA method, rivaling even the most advanced ones in terms of performance.

\subsubsection{Source HypOthesis Transfer}\label{sub:shot}
Source HypOthesis Transfer (\textbf{SHOT})~\citep{liang2020we} has become a prominent reference in the field of SF-UDA. 
Its efficacy has set a baseline for newer methodologies in the field. 
The adaptation in SHOT is characterized by a sequence of epochs, each consisting of two phases:

\begin{enumerate}
    \item \textbf{Pseudo-labels generation}: label the target samples through the source-trained model.
    \item \textbf{Feature extractor adaptation}: tune the feature extractor weights optimizing an objective composed of two terms: the unsupervised \emph{Information Maximization loss} and a standard \emph{Cross-Entropy loss} based on the previously computed pseudo-labels.
    Notably, the classifier weights remain unaltered in this phase.
\end{enumerate}

\noindent For pseudo-labels computation, SHOT employs a modified version of SCA. Utilizing the feature extractor $f(\cdot)$ and the fixed source classifier, the algorithm determines class output probabilities for each target sample \( \mathbf{x}^{(i)} \in \mathcal{X}_{\text{TGT}}\) as \( \mathbf{p}^{(i)} = (p_{1}^{(i)}, \ldots, p_{C}^{(i)}) \), where $C$ is the total number of classes. The initial prototype $\mathbf{k}_c$ for a generic class $c \in \{1, \ldots, C\}$, is obtained as a weighted average of target domain features, according to the following equation:

\begin{equation}
    \mathbf{k}_c = \frac{\sum_i p_c^{(i)} \cdot f(\mathbf{x}^{(i)})}{\sum_i p_c^{(i)}},
\end{equation}

\noindent Where $i$ indexes all the samples of the target domain $\mathcal{X}_{\text{TGT}}$. The obtained prototypes \(\{\mathbf{k}_c\}_{c=1}^{C}\) serve as the initialization for the spherical K-Means steps of SCA. This leads to a 1-NN classifier generating the pseudo-labels for the target dataset.

\subsubsection{Neighborhood Reciprocity Clustering}\label{sub:nrc}
Neighborhood Reciprocity Clustering (\textbf{NRC})~\citep{yang2021exploiting} leverages the intrinsic latent structure of target data by focusing on neighborhood information.
Specifically, the training aims to bring similar samples closer while distancing dissimilar ones. NRC builds on the concept of \textbf{Neighborhood Affinity} distinguishing neighbors by their potential use for accurate supervision.
Neighbors are categorized as either reciprocal (RNN) or non-reciprocal (nRNN). RNNs are neighbors that mutually identify each other as nearest neighbors and so, provide valuable adaptation information.
Consequently, more importance is assigned to RNNs in the overall objective.
This perspective is further extended by the \textbf{Expanded Neighborhood Affinity}, which considers second-order neighbors (i.e., neighbors of neighbors).
NRC devises an objective by integrating these components, ensuring a balanced consideration of both immediate and extended neighborhoods. 
Finally, a \textbf{Self-Regularization} term employs current predictions as a guide to mitigate the influence of potentially unreliable neighbors.

\subsubsection{Attracting and Dispersing}\label{sub:aad}

Attracting and Dispersing (\textbf{AAD})~\citep{yang2022attracting} bears similarities to NRC, but introduces a simpler methodology and objective.
For every target feature vector, two distinct sets are determined: the \textbf{Close Neighbor Set}, which encompasses the nearest or most similar representations, and the \textbf{Background Set}, consisting of feature vectors unrelated or distant from the current one (often from different classes). 
The method employs the principles of \textbf{Attracting} and \textbf{Dispersing}. In \textit{Attracting}, the Close Neighbor Set's representations are guided to produce consistent predictions.
Conversely, in \textit{Dispersing}, predictions from the Background Set are kept as distant as possible from the ones obtained with the current representations. The model's parameters are optimized with a simple objective to achieve a balance between attraction and dispersion, iteratively refining the process until the model adapts to the target domain.

\subsubsection{Polycentric Clustering and Structural Regularization}\label{sub:pcsr}
Polycentric Clustering and Structural Regularization (\textbf{PCSR})~\citep{guan2022polycentric} is similar to SHOT since it employs the Information Maximization loss and a pseudo-labeling procedure. However, PCSR refines the pseudo-labeling through \textbf{Polycentric Clustering}. Leveraging on the K-Means algorithm, it computes multiple centers for each class, enhancing pseudo-label accuracy.
PCSR introduces also the use of \textbf{MixUp}~\citep{zhang2018mixup, guo2019mixup, carratino2022mixup} to provide additional regularization during the adaptation.
The result is a composite loss function integrating all the aforementioned objectives for the unsupervised model optimization on the target domain.

%% file: Framework.tex
A significant contribution of this work is the introduction of an open-source benchmarking framework enabling the execution of a systematic large-scale experimental analysis of SF-UDA.
Our framework facilitates the construction, training, and testing of SF-UDA methods. 
Thanks to the integration with  state-of-the-art libraries, e.g., \texttt{timm}~\citep{rw2019timm}, it enables comparisons between more than 500 backbones.
Furthermore, it allows to test SF-UDA methods on several datasets, ranging from widely used benchmarks, such as \textit{Office 31}~\citep{saenko2010adapting} and \textit{Office-Home}~\citep{venkateswara2017deep}, to other more recent datasets, such as \textit{DomainNet}~\citep{peng2019moment}. 
\noindent Importantly, the  proposed framework also allows to modify and assess individual components of distinct methods in a flexible way.
This enables three primary outcomes:
\begin{enumerate}
    \item A deeper understanding of the SF-UDA setting, which is currently understudied;
    \item The establishment of a systematic protocol for the comparison of different SF-UDA methods;
    \item Guidance in making key choices during the design of SF-UDA approaches.
\end{enumerate}

In the following sections, we present the datasets (Sec.~\ref{subsec:datasets}) and the backbones (Sec.~\ref{subsec:backbones}) considered in our study. Furthermore, we give details on the benchmark protocol in Sec.~\ref{subsec:protocol}.

\subsection{Datasets}\label{subsec:datasets}
We now introduce all datasets considered in our study, distinguishing between those used for pre-training and for domain adaptation.

\noindent\textbf{Pre-training datasets.} For the \textit{first transfer}, we consider two datasets: ImageNet, composed of 1.2\,M images for $1\,000$ mutually exclusive classes, and the superset ImageNet21k~\citep{deng2009imagenet}, composed of 14\,M images for $21\,841$ not mutually exclusive classes.
Specifically, we either consider models pre-trained (i) on ImageNet (\textit{IN}), or (ii) on ImageNet21k and consequently fine-tuned on the $1\,000$ classes of ImageNet (\textit{IN21k}).\\
\noindent\textbf{Domain adaptation image datasets.} These image classification datasets include two or more subsets corresponding to different domains with distinct visual styles and  sharing the same classes.
In our experiments, we consider:
\textit{DomainNet}~\citep{peng2019moment}, \textit{ImageClef-DA}~\citep{long2017deep}, \textit{Office-31}~\citep{saenko2010adapting}, \textit{Modern Office-31}~\citep{ringwald2021adaptiope}, \textit{Visda-2017}~\citep{peng2017visda}, \textit{Office-Home}~\citep{venkateswara2017deep}, and \textit{Adaptiope}~\citep{ringwald2021adaptiope}, 
An overview of all datasets is illustrated in Tab.~\ref{table:datasets}.

\noindent\textbf{Datasets choice rationale.} For pre-training, we include ImageNet and ImageNet21k, as they are open datasets frequently employed in the literature.
Consequently, many DNNs pre-trained on these datasets are available.
Instead, for domain adaptation, we select 7 datasets.
These include popular benchmark datasets (Office31, Office-Home, and Visda) and also other datasets less often used in SF-UDA research (like DomainNet, Adaptiope, Modern Office-31, and ImageClef-DA).
The goal of this diverse selection is to ensure a comprehensive evaluation across different scenarios, supporting our findings regarding SF-UDA.
To evaluate different SF-UDA methods (Sec.~\ref{sec:methods}), we select Office31 and Office-Home, as they are used as benchmarks in the original papers, and Adaptiope and ImageClef-DA, which are not considered in those studies.
This allows to determine if the methods can generalize to new datasets or if their performance strongly depends on the dataset choice.
For the experiments on pre-training and fine-tuning (Sec.~\ref{sec:pretraining} and~\ref{sec:finetuning}), we employ all the domain adaptation datasets mentioned above with the exception of Office31 due to its similarity to Modern Office-31.
This results in a total of 23 domains and 74 domain pairs (source-target). 
Notably, for Visda-2017, unlike the usual benchmark, we consider both synthetic-to-real and real-to-synthetic experiments.
We run an experiment for each domain pair (a source and target domain) and we average the accuracy across experiments.

\begin{table*}
    \centering
    \caption{Domain Adaptation datasets considered in this study.
    Checkmarks in the final two columns show which datasets are used in the SF-UDA methods evaluation and which in the pre-training and fine-tuning (PT/FT) experiments.} \label{table:datasets}
    \begin{tabular}{lccccc}
    \textbf{Dataset} & \textbf{Images} & \textbf{Domains}  & \textbf{Classes} & \textbf{SF-UDA Eval}  & \textbf{PT/FT Exp.} \\
    \midrule
    DomainNet & $596\,006$ & $6$  & $345$ &  & $\checkmark$ \\
    Visda-2017 & $280\,000$ & $2$  & $12$ &  & $\checkmark$ \\
    Adaptiope & $36\,900$ & $3$  & $123$ & $\checkmark$ & $\checkmark$ \\
    Office-Home & $15\,588$ & $4$ & $65$ & $\checkmark$ & $\checkmark$ \\
    Modern Office-31 & $7\,210$ & $4$  & $31$ &  & $\checkmark$  \\
    Office-31 & $4\,110$ & $3$  & $31$ & $\checkmark$ &  \\
    ImageClef-DA & $2\,400$ & $4$  & $12$ &  $\checkmark$ & $\checkmark$ \\
    \bottomrule
    \end{tabular}

\end{table*}

\subsection{Backbone selection and integration}\label{subsec:backbones}
Before any SF-UDA pipeline, the backbone should be chosen (see Sec.~\ref{subsec:sfuda_pipeline}): in most of our experiments we rely on the \textit{PyTorch Image Models} Python library (\texttt{timm})~\citep{rw2019timm} that provides access to a remarkably large number of different architectures and pre-trained weights. 
Specifically, to analyze the pre-training phase, in Sec.~\ref{sec:pretraining}  we evaluate SCA and the baselines (presented in Sec.~\ref{subsec:protocol})  on more than 500 models selected among more than 25 different families of architectures (e.g., VGG~\citep{vgg}, ResNet~\citep{resnet}, EfficientNet~\citep{efficientnet}, ConvNeXt~\citep{convnext}, ViT~\citep{dosovitskiy2020image}, SWIN~\citep{swin}, Deit~\citep{deit} and XCiT~\citep{xcit}). Moreover, to evaluate the impact of the fine-tuning step on the source domain (Sec.~\ref{sec:finetuning} and some experiments of Sec.~\ref{sec:pretraining}), we sample a subset of 59 models, taken from more than 12 families of architectures. 
Instead, to benchmark SF-UDA methods and test their robustness (Sec.~\ref{sec:methods}), we consider two commonly used backbones, namely the convolution-based Resnet50~\citep{resnet} (that we initialized with TorchVision\footnote{\url{https://github.com/pytorch/vision}} weights for consistency with the original papers) and the Transformer-based ViT-Large~\citep{dosovitskiy2020image}.
Finally, for experiments involving self-supervised pre-trained networks (Sec.~\ref{subsec:ssl}), we use the weights released by the authors that proposed the different methods if not available on TorchVision Hub\footnote{\url{https://pytorch.org/hub/}}.
Notably, our analysis comprises both modern Vision Transfomers and CNNs (like ConvNeXt) and more traditional architectures.

\subsection{Evaluation protocol}\label{subsec:protocol}

\begin{table*}[t!]
\caption{Evaluation baselines and SF-UDA settings considered in this work. In the first section on the left the data of supervised training (\textbf{First Transfer}) is specified (i.e. the \textbf{Domain}) together with the training strategy: optimise only the weight of a classifier or fine-tune also the backbone. The following columns indicate the adaptation domain available for the \textbf{Second transfer} and the domain using to test the performance.}

\label{tab:tasks}
\begin{tabular*}{\textwidth}{@{\extracolsep\fill}lcccc}
\toprule%
\textbf{\multirow{2}{*}{Task}} & 
\multicolumn{2}{@{}c@{}}{ \textbf{First Transfer (supervised)} }  & 
\textbf{\multirow{2}{*}{\shortstack{Second Transfer \\ Domain}}} & 
\textbf{\multirow{2}{*}{\shortstack{Test \\ Domain}}} \\

& \textbf{Domain} & \textbf{Optimization} & & \\
\midrule
\textbf{LP-IDG}   & Target (w/ labels) & Classifier only & None & Target \\
\textbf{FT-IDG}   & Target (w/ labels) & Backbone+Classifier  & None & Target \\
\hdashline\noalign{\vskip 0.7ex}
\textbf{LP-ODG}   & Source (w/ labels) & Classifier only & None & Target \\
\textbf{FT-ODG}   & Source (w/ labels) & Backbone+Classifier & None & Target \\
\hdashline\noalign{\vskip 0.7ex}
\textbf{SF-UDA}    & Source (w/ labels) & Classifier only & Target (w/o labels) & Target \\
\textbf{FT-SF-UDA} & Source (w/ labels) & Backbone+Classifier  & Target (w/o labels) & Target \\

\botrule
\end{tabular*}

\end{table*}

Our experimental framework is designed to serve as a toolkit for the in-depth analysis of SF-UDA and it enables the evaluation of each individual component of the SF-UDA pipeline.
Additionally, our evaluation protocol is standardized across datasets and architectures in order to provide reliable and reproducible assessments. 

\noindent \textbf{SF-UDA protocol}. For every pre-trained backbone, the final linear layer (i.e., the ImageNet classifier) is removed.
Then, we introduce a newly initialized bottleneck, followed by a final linear classification layer tailored to the number of classes of the given task.
The bottleneck includes a linear layer, which projects the backbone's features to $256$ dimensions, followed by a normalization layer and a nonlinear activation function. Notably, the type of normalization adopted for the bottleneck is contingent on the backbone's architecture, opting for Batch Normalization if present in the backbone or Layer Normalization otherwise. Similarly, the activation function can be a ReLU or GELU based on the backbone's activations.

If the training on the \textit{source domain} is performed with fine-tuning (FT), backbone weights are adjusted and the bottleneck and classifier weights are learnt from scratch. Instead, experiments with no FT keep the backbone weights fixed and train the bottleneck and the classifier only.
Finally, the adaptation on the \textit{target domain} is performed with the selected SF-UDA technique.

\noindent \textbf{Experimental baselines}. We also implement four baselines
serving as upper and lower bounds for SF-UDA methods.
For these baselines, we consider the following two \textit{data settings}:

\begin{itemize}
\item \textbf{In-Domain Generalization} (\textbf{IDG}). This represents the standard ``generalization" typically considered in supervised learning. In this context, the network is trained and tested on a single labeled domain that, for consistency, we call \textit{target}. Images from this domain are randomly partitioned into a training set (80\%) and a test set (20\%).
We regard IDG as the upper bound for SF-UDA.
\item \textbf{Out-of-Domain Generalization} (\textbf{ODG}). In this setting, the pre-trained network is trained on the (labeled) \textit{source domain} and tested on the \textit{target domain}. Given that no adaptation takes place, we consider this as the lower bound for SF-UDA.
\end{itemize}

\noindent Moreover, we consider the following two baseline \textit{training strategies}:

\begin{itemize}
\item \textbf{Linear Probing} (\textbf{LP}). The pre-trained backbone weights are kept frozen and only a linear classifier is trained.
\item \textbf{Fine-Tuning} (\textbf{FT}). We stack a bottleneck and a new linear classifier on top of the backbone and we fine-tune end-to-end the entire model.
\end{itemize}

\noindent As a result, we end up with four baselines: \textbf{LP-IDG} and \textbf{LP-ODG} for in-domain and out-of-domain generalization with linear probing and \textbf{FT-IDG} and \textbf{FT-ODG} for in-domain and out-of-domain generalization with fine-tuning. See Tab.~\ref{tab:tasks} for a comparative summary of the baselines and the considered SF-UDA settings. 

In our study, we adhere to the transductive setting~\citep{kouw2019review}: the final target accuracy is evaluated using the same images that were used during the unsupervised adaptation process, aligning with standard practices in UDA and SF-UDA.
We evaluate the performance of various methodologies based on classification accuracy and failure rate on the target domain.
In particular, we consider a SF-UDA method to fail an experiment if the final target accuracy is lower than the LP-ODG baseline. Detailed insights on accuracy variations of the considered methods can be found in Sec.~\ref{sec:finetuning}.

%% file: experiments_methods.tex
In this section we experiment with SCA, SHOT, NRC, AAD, and PCSR, as motivated in Sec.~\ref{subsec:sfuda_methods}, highlighting both the strengths and weaknesses of each method across different settings. Note that, as analyzed in Sec.~\ref{sec:finetuning}, fine-tuning can degrade performance with Batch Normalization. For this reason, as SCA aligns only the classifier to the new domain (without adapting the backbone and the Batch Normalization statistics), it does not properly work in this unfavourable setting. Thus, for a fair comparison we did not perform any source fine-tuning when using ResNet50 for SCA.
Our evaluation centers on the following research questions.

\noindent\textbf{Reproducibility}. Can the selected SF-UDA methods be consistently reproduced in more general experimental settings with respect to those presented in their original evaluations?

\noindent\textbf{Distributed training}. SF-UDA methods are often tested on small and medium size datasets using backbones such as ResNet50 (not computationally demanding). 
However, it is important to study their ability to distribute computation across GPUs, especially when large-scale datasets are used or when modern, more computationally demanding backbones (e.g., Vision Transformers and ConvNeXt) are adopted.
    
\noindent\textbf{Generalization across datasets}. Can SF-UDA methods generalize to datasets not  considered in their original papers?

We aim at identifying biases and better characterize SF-UDA methods' performances in more general scenarios.

\noindent\textbf{Backbone independence}.  While the original papers show how these methods can be beneficial on specific backbones (such as ResNet50 or ResNet101), can they provide performance gains also with more recent and powerful architectures (such as Vision Transformers)?
    
\noindent\textbf{Hyperparameter sensitivity}. How sensitive are these methods to hyperparameter tuning?
Since in SF-UDA target domain labels are absent during adaptation, hyperparameter selection becomes challenging.
This may lead to fundamental issues such as data leakage (selecting hyperparameters based on target test accuracy after trial and error), which can misrepresent performance metrics.
Assessing robustness to hyperparameter changes can shed light on a method's real-world applicability.\\

\subsection{Reproducibility}
\label{sub:reproducibility}
In this section, we seek to identify methods maintaining consistent performance across diverse settings and robust to variations of the random seed and software version.
To this aim, we set up a testing environment, distinct from those used in the original papers code.
We conduct the experiments using five different random seeds, a departure from the prevalent practice in UDA and SF-UDA literature, which typically reports results from a single run.
As evidenced in Tab.~\ref{tab:comparison_accuracies}, most methods demonstrate robustness to our testing conditions, although AAD and NRC, on average, perform marginally below the results reported in the reference papers.
Importantly, performance variations observed in our experiments cast doubts on the recurring practice of reporting single-seed performances in \textit{benchmark tables}.
Using such results to rank methods is questionable, as the disparities between them are often smaller than the standard deviations we observed.
Thus, relying solely on results from a single run fails to provide a significant evaluation.
SCA is not included in the table, since we are the first presenting and studying it as a stand-alone SF-UDA method.

\input{reproducibility_table}

\input{methods_table}

\subsection{Distributed Training}
\label{sub:distributed}
While distributed training enables computational scalability, its potential remains largely unexplored in SF-UDA research.
This can be attributed to the prevalent practice of benchmarking methods using architectures such as ResNet50 and ResNet101, which are not computational demanding according to current standards.
However, as we discuss in section \ref{sec:pretraining}, the choice of architecture and pre-training strategy can profoundly influence outcomes. 
In fact, moving from ResNet50 to recent architectures, such as ViT or ConvNeXt, might offer significant enhancements to target accuracy. For example, the basic FT-ODG performance of the ViT-Large model significantly outperforms the ResNet50 when applying any SF-UDA method, as detailed in Table~\ref{tab:average_results}.
This underscores the importance of effective distributed computations for SF-UDA methods.

\noindent SCA performance is not affected by the distribution of the computations. This is attributed to the adaptation of the classifier using only K-Means, which eliminates the need for network optimization. Both feature extraction and K-Means updates can be effectively performed in parallel on multiple GPUs without affecting the final outcome, ensuring the accuracy of the final results remains consistent, irrespective of the number of GPUs utilized.
Instead, most SF-UDA methods (including all the others considered in this work) incorporate a diversity term~\citep{yang2022attracting} in their objective, whose exact computation may require the entirety of the target dataset at once~\citep{liang2020we}.
In practice, the diversity term is often approximated using the current batch during the training process.
When this batch is split across multiple GPUs and the gradients are averaged (a common practice in distributed training), it often results in a less accurate approximation than the centralized one.
Hence, performance degradation may occur when SF-UDA methods are distributed.

\noindent We empirically analyze this matter in Tab.~\ref{table:methods_table}, reporting the performance
with the standard global batch size of 64 distributed across 1, 2, 4, 8, and 16 GPUs. 
The table shows different results for different methods.
Specifically, SHOT and PCSR exhibit robustness, allowing for efficient parallelization. Conversely, AAD and NRC are more sensitive, necessitating larger batch sizes for effective performance. Smaller batch sizes can, at times, entirely degrade their efficacy, especially with ResNet50 for which the performance drop is up to~55\%. We observe the same behaviour across all datasets.

\subsection{Dataset and architecture independence}
\label{sub:dataset_architecture}
The datasets frequently employed in SF-UDA experimental analyses typically have small to medium sizes or  include only few domains.
This can lead to involuntary data leakage and unreliable performance evaluations of the methods.
For instance, iterative design choices and hyperparameter tuning, guided by observations of target accuracy (trial and error procedure), can inadvertently cause a method to specialize excessively to a particular benchmark dataset with the selected backbone.
This can potentially mislead researchers into considering a method as universally effective while it may be excessively tailored (i.e., \textit{overfitted}) to specific datasets.
Furthermore, the architecture choice can influence SF-UDA performance and, even though ResNets are the standard choice in experimental evaluations, it is crucial to establish whether the methods are robust to backbone changes. To address these questions, our experimental framework includes two datasets not commonly featured in SF-UDA papers (i.e., Adaptiope and Image-CLEF) together with two datasets that, instead, are normally considered (i.e., Office31 and Office-Home) as reported in Tab~\ref{table:datasets}. 
Moreover, we use ViT-Large to evaluate the methods beyond well-studied ResNet backbones.

The selection of ViT-Large is driven by the following factors: it offers a distinct inductive bias compared to ResNets, characterized by its structure as a Vision Transformer incorporating Self-Attention and Layer Normalization, unlike the Convolutional Neural Network with Batch Normalization Layers seen in ResNets. 
Additionally, ViT has demonstrated enhanced generalization capabilities on ImageNet and many other Computer Vision tasks.
Note that the aim of this experiment is to test the robustness of the different SF-UDA methods, while a comprehensive study on the impact of the backbone and pre-training choices is reported in Sec.~\ref{sec:pretraining}.

We report our results in Tab.~\ref{table:methods_table} and Tab.~\ref{tab:average_results}.
In summary, SHOT and PCSR demonstrate robustness across different architectures and datasets. NRC exhibits competitive performance with different datasets when ResNet50 is employed, but it shows sub-optimal performance with ViT.
Moreover, AAD's performance is negatively affected be changes in either the dataset, architecture, or both.
It is worth noting that while a higher accuracy is attainable with ViT, all methods display larger standard deviations, due to possible failures or sub-optimal adaptation.
Finally, SCA demonstrates strong performance when ViT is employed, while it achieves sub-optimal performance with ResNet50. 
As we will see in Sec.~\ref{sec:finetuning}, this is due to the failure in some domain-pairs caused by Batch Normalization Layers: since SCA aligns just the classifier without optimizing the backbone weights, large differences in the BN statistics can cause the algorithm to fail.

\noindent \textbf{Remark.} The under-performance of NRC and AAD on ViT-Large, even falling below the FT-ODG baseline (i.e., fine-tuning without any adaptation), despite ViT's superior inductive bias and stronger out-of-distribution generalization (as indicated in Table~\ref{tab:average_results}), suggests that while these methods might be effective on architectures different from ResNets, they likely require modifications to be compatible with more advanced architectures like Vision Transformers.

\input{methods_average_results_table}

\subsection{Hyperparameter Sensitivity} \label{sub:hparams}

In practical applications of SF-UDA, determining the optimal hyperparameters can be challenging without incurring in data leakage. Parameters like the number of training epochs and learning rate are less problematic. They can be adjusted based on the optimization of the unsupervised objective during adaptation. Many SF-UDA methods default to the same learning rate, schedule, weight decay, optimizer, etc. However, SF-UDA algorithms may introduce specific hyperparameters sensitive to particular experimental conditions.

Unsupervised hyperparameter selection techniques, as suggested by \cite{saito2021tune}, could be a solution. Yet, their effectiveness, particularly their robustness to architecture and dataset variations, requires further study.

While SCA is hyperparameters-free, methods like SHOT and PCSR present some hyperparameters, described in their original papers, that are never changed and we treat them as universal constants: despite being likely found on a specific dataset, their effective application in varied contexts (as shown in the original reports) suggest they can be used effectively across datasets and architectures without modifications. Conversely, methods like AAD and NRC have dataset-specific hyperparameters that may be hard to tune in practice. Our analysis focuses on the hyperparameters $K$ and $\beta$ for AAD (number of neighbors and a loss-scheduling parameter, respectively) and $K$ and $KK$ for NRC (number of neighbors and second-order neighbors, respectively). For detailed information on these hyperparameters, refer to the original publications and code.

Tables~\ref{tab:aad_hparams} and \ref{tab:nrc_hparams} show average accuracy values across four datasets (Office31, Office-Home, Adaptiope, and Image-CLEF), using hyperparameters from the respective original papers. AAD's average performance falls behind NRC's, which aligns more closely with SHOT and PCSR (with average accuracies of $76.34\%$ and $77.45\%$, respectively, on ResNet50). A significant challenge with NRC is observed in its high accuracy on the VisDA dataset using ResNet101 (achieving a macro-averaged accuracy of $85.90\%$) with $K=5$ and $KK=5$. However, as our tables indicate, these settings do not perform well on the datasets in our experiments.

\begin{table}[h!]
    \centering
    \setlength{\tabcolsep}{-4pt}
    \captionsetup{width=0.9\linewidth}
    \caption{Hyperparamter test for AAD with ResNet50 (mean accuracy on Office31, Office Home, Adaptiope and Image-Clef datasets).} \label{tab:aad_hparams}
   
    \begin{tabular}[t]{l|cc}
            \textbf{AAD} & \textbf{K=3} & \textbf{K=5} \\
            \midrule
            \boldmath$\beta=0$ \newtabsection \gradientsingle{aad_hparams}{75.70} & \gradientsingle{aad_hparams}{73.16} \n
            \boldmath$\beta=0.75$ \newtabsection \gradientsingle{aad_hparams}{76.34} & \gradientsingle{aad_hparams}{73.28} \n
            \boldmath$\beta=1$ \newtabsection \gradientsingle{aad_hparams}{76.20} & \gradientsingle{aad_hparams}{73.03} \n
            \boldmath$\beta=2$ \newtabsection \gradientsingle{aad_hparams}{75.36} & \gradientsingle{aad_hparams}{71.83} \n
            \boldmath$\beta=5$ \newtabsection \gradientsingle{aad_hparams}{73.28} & \gradientsingle{aad_hparams}{69.52} \n
            \bottomrule
    \end{tabular}
\end{table}

\begin{table}[h!]
    \centering
    \setlength{\tabcolsep}{-4pt}
    \captionsetup{width=0.9\linewidth}
    \caption{Hyperparamter test for NRC on ResNet50 (mean accuracy on Office31, Office Home, Adaptiope and Image-Clef datasets).} \label{tab:nrc_hparams}
      
        \begin{tabular}[t]{l|cccc}
            \textbf{NRC} \newtabsection \textbf{K=2} & \textbf{K=3} & \textbf{K=4} & \textbf{K=5} \\
            \midrule
            \textbf{KK=2} \newtabsection \gradientsingle{nrc_hparams}{75.37} & \gradientsingle{nrc_hparams}{77.03} & \gradientsingle{nrc_hparams}{77.39} & \gradientsingle{nrc_hparams}{77.28} \n
            \textbf{KK=3} \newtabsection \gradientsingle{nrc_hparams}{76.87} & \gradientsingle{nrc_hparams}{77.45} & \gradientsingle{nrc_hparams}{77.19} & \gradientsingle{nrc_hparams}{76.56} \n
            \textbf{KK=4} \newtabsection \gradientsingle{nrc_hparams}{77.31} & \gradientsingle{nrc_hparams}{77.31} & \gradientsingle{nrc_hparams}{76.47} & \gradientsingle{nrc_hparams}{75.52} \n
            \textbf{KK=5} \newtabsection \gradientsingle{nrc_hparams}{77.40} & \gradientsingle{nrc_hparams}{76.78} & \gradientsingle{nrc_hparams}{75.57} & \gradientsingle{nrc_hparams}{74.33} \n
            \bottomrule
        \end{tabular}

\end{table}

\input{summary_table}

\subsection{Remarks}
\label{sub:remarks}
The results discussed in previous sections reveal insights on the reproducibility, possibility of distributed training, dataset, backbone, and hyperparameter variations robustness on the considered SF-UDA methods.
We summarize our findings in Tables~\ref{tab:average_results} and \ref{tab:summary}.
In the next sections, we focus on SCA and SHOT and exclude AAD and NRC due to previously discussed limitations. Given the similarity of PCSR to SHOT in terms of adaptation performance and design choices, in the following we consider SHOT for its simplicity. Further, SHOT and SCA are complementary: while SHOT, during adaptation, modifies the weights of the backbone keeping the classifier fixed, SCA only adapts the classifier without tuning the backbone weights. The issues of SCA with ResNet50 are discussed and analyzed in Sec.~\ref{sec:finetuning}, where we will see that they are caused by BN layers.

%% file: reproducibility_table.tex
\begin{table}[h!]
    \centering
    \caption{Accuracy comparison between paper results and our findings on \textit{Office31} and \textit{Office-Home} datasets. We report the mean accuracy and the standard deviation of 5 runs.} \label{tab:comparison_accuracies}
    \begin{tabular}{c C{1cm}C{1cm}|C{1cm}C{1cm}}

    & \multicolumn{2}{c|}{\textbf{OFFICE31}} & \multicolumn{2}{c}{\textbf{OFFICE-HOME}} \\
    \cmidrule{2-5}
    \textbf{} & \textbf{Reported} & \textbf{Ours} & \textbf{Reported} & \textbf{Ours} \\
    \midrule
    \textbf{AAD} & $89.9$ & $88.8 \mathsmaller{\pm 0.5}$ & $72.7$ & $71.4 \mathsmaller{\pm 0.3}$ \\
    \midrule
    \textbf{NRC} & $89.4$ & $88.4 \mathsmaller{\pm 0.6}$ & $72.2$ & $70.8 \mathsmaller{\pm 0.1}$ \\
    \midrule
    \textbf{SHOT} & $88.6$ & $88.5 \mathsmaller{\pm 0.3}$ & $71.8$ & $72.1 \mathsmaller{\pm 0.2}$ \\
    \midrule
    \textbf{PCSR} & $89.5$ & $89.4 \mathsmaller{\pm 0.5}$ & $72.8$ & $72.7 \mathsmaller{\pm 0.3}$ \\
    \bottomrule
    \end{tabular}

\end{table}

%% file: methods_table.tex
\begin{table*}[!htbp]
\caption{Evaluation of SF-UDA methods with ResNet50 and ViT-Large on \textbf{Office31}, \textbf{Office-Home}, \textbf{Adaptiope} and \textbf{Image-CLEF} datasets (for each experiment mean and standard deviation are evaluated over 5 runs). The \textit{global batch size} is fixed to $64$. Different rows represent different distributed setting (\#GPUs $\times$ local batch size).} \label{table:methods_table}

    
    \setlength{\extrarowheight}{7.2pt}
    \setlength{\aboverulesep}{0pt}
    \setlength{\belowrulesep}{3pt}

    \renewcommand\cellset{\renewcommand\arraystretch{0.45}}
    \setlength{\tabcolsep}{-4pt}

    \centering
    \begin{tabular}{p{0.8cm}c*{5}{c}|*{5}{c}}
        & & \multicolumn{5}{c}{\textbf{ResNet50}} & \multicolumn{5}{c}{\textbf{ViT-Large}} \n
        \cmidrule{3-12} 
        & &  \textbf{ SCA} & \textbf{AAD} & \textbf{NRC} & \textbf{SHOT} & \textbf{PCSR} & \textbf{ SCA} & \textbf{AAD} & \textbf{NRC} & \textbf{SHOT} & \textbf{PCSR} \n
        \midrule
        
        \multirow{5}{*}{\rotatebox[origin=b]{90}{\textbf{OFFICE 31 \hspace{0.2cm}}} } & \textbf{1x64} \newtabsection 
        \multirow{5}{*}{\gradientfull{office31_resnet50}{84.5}{0.4}} &
        \gradient{office31_resnet50}{88.8}{0.5} & \gradient{office31_resnet50}{88.4}{0.6} & \gradient{office31_resnet50}{88.5}{0.3} & \gradient{office31_resnet50}{89.4}{0.5} \newtabsection \multirow{5}{*}{\gradientfull{office31_vit}{94.9}{0.2}} & \gradient{office31_vit}{94.4}{0.5} & \gradient{office31_vit}{94.9}{0.2} & \gradient{office31_vit}{94.3}{0.4} & \gradient{office31_vit}{94.1}{1.8} \n
        
        & \textbf{2x32} \newtabsection
        &
        \gradient{office31_resnet50}{86.8}{0.7} & \gradient{office31_resnet50}{87.4}{0.5} & \gradient{office31_resnet50}{88.5}{0.6} & \gradient{office31_resnet50}{89.4}{0.5} \newtabsection & \gradient{office31_vit}{94.2}{0.5} & \gradient{office31_vit}{93.4}{1.2} & \gradient{office31_vit}{94.2}{0.3} & \gradient{office31_vit}{94.3}{0.7} \n
        
        & \textbf{4x16} \newtabsection 
        &
        \gradient{office31_resnet50}{83.0}{0.9} & \gradient{office31_resnet50}{84.4}{0.4} & \gradient{office31_resnet50}{88.4}{0.4} & \gradient{office31_resnet50}{89.3}{0.4} \newtabsection & \gradient{office31_vit}{93.0}{0.9} & \gradient{office31_vit}{90.1}{3.0} & \gradient{office31_vit}{94.4}{0.2} & \gradient{office31_vit}{94.7}{0.3} \n
        
        & \textbf{8x8}  \newtabsection 
        & \gradient{office31_resnet50}{76.8}{1.1} & \gradient{office31_resnet50}{77.1}{0.3} & \gradient{office31_resnet50}{87.6}{0.2} & \gradient{office31_resnet50}{89.1}{0.5} \newtabsection 
        & \gradient{office31_vit}{91.8}{0.5} & \gradient{office31_vit}{75.5}{0.8} & \gradient{office31_vit}{94.5}{0.3} & \gradient{office31_vit}{93.9}{1.7} \n
        
        & \textbf{16x4} \newtabsection 
        & \gradient{office31_resnet50}{57.9}{3.0} & \gradient{office31_resnet50}{62.4}{1.5} & \gradient{office31_resnet50}{86.8}{0.4} & \gradient{office31_resnet50}{88.1}{0.7} \newtabsection 
        &\gradient{office31_vit}{86.4}{5.6} & \gradient{office31_vit}{52.9}{1.3} & \gradient{office31_vit}{94.4}{0.2} & \gradient{office31_vit}{94.0}{1.7} \n
        \midrule
        
        \multirow{5}{*}{\rotatebox[origin=b]{90}{\textbf{OFFICE-HOME \hspace{0.1cm}}}} & \textbf{1x64} \newtabsection 
        \multirow{5}{*}{\gradientfull{officehome_resnet50}{66.1}{0.1}} &
        \gradient{officehome_resnet50}{71.4}{0.3} & \gradient{officehome_resnet50}{70.8}{0.1} & \gradient{officehome_resnet50}{72.1}{0.2} & \gradient{officehome_resnet50}{72.7}{0.3} \newtabsection
        \multirow{5}{*}{\gradientfull{officehome_vit}{87.2}{0.2}} & \gradient{officehome_vit}{80.9}{4.4} & \gradient{officehome_vit}{81.0}{2.4} & \gradient{officehome_vit}{88.0}{0.9} & \gradient{officehome_vit}{85.1}{3.9} \n
        
        & \textbf{2x32} \newtabsection & \gradient{officehome_resnet50}{67.4}{0.3} & \gradient{officehome_resnet50}{67.7}{0.2} & \gradient{officehome_resnet50}{71.9}{0.2} & \gradient{officehome_resnet50}{72.3}{0.6} \newtabsection & \gradient{officehome_vit}{80.8}{5.3} & \gradient{officehome_vit}{77.8}{5.4} & \gradient{officehome_vit}{86.5}{3.1} & \gradient{officehome_vit}{84.6}{2.6} \n
        
        & \textbf{4x16} \newtabsection & \gradient{officehome_resnet50}{60.5}{0.5} & \gradient{officehome_resnet50}{51.9}{0.4} & \gradient{officehome_resnet50}{71.7}{0.2} & \gradient{officehome_resnet50}{72.6}{0.2} \newtabsection &     
        \gradient{officehome_vit}{75.0}{7.2} & \gradient{officehome_vit}{68.7}{2.2} & \gradient{officehome_vit}{87.7}{0.7} & \gradient{officehome_vit}{85.3}{3.2} \n
        
        & \textbf{8x8}  \newtabsection & \gradient{officehome_resnet50}{49.1}{0.4} & \gradient{officehome_resnet50}{22.4}{0.5} & \gradient{officehome_resnet50}{71.1}{0.2} & \gradient{officehome_resnet50}{72.0}{0.2} \newtabsection & \gradient{officehome_vit}{76.0}{5.6} & \gradient{officehome_vit}{56.4}{2.4} & \gradient{officehome_vit}{86.3}{3.6} & \gradient{officehome_vit}{81.9}{5.8} \n
        
        & \textbf{16x4} \newtabsection & \gradient{officehome_resnet50}{22.8}{0.7} & \gradient{officehome_resnet50}{31.7}{0.6} & \gradient{officehome_resnet50}{68.9}{0.3} & \gradient{officehome_resnet50}{70.2}{0.1} \newtabsection & \gradient{officehome_vit}{70.2}{12.2} & \gradient{officehome_vit}{46.9}{1.6} & \gradient{officehome_vit}{86.5}{1.3} & \gradient{officehome_vit}{85.9}{3.8} \n
        \midrule
        
        \multirow{5}{*}{\rotatebox[origin=b]{90}{\textbf{ADAPTIOPE \hspace{0.2cm}}}} & \textbf{1x64} \newtabsection \multirow{5}{*}{\gradientfull{adaptiope_resnet50}{42.1}{0.7}} & \gradient{adaptiope_resnet50}{59.9}{1.5} & \gradient{adaptiope_resnet50}{64.9}{0.6} & \gradient{adaptiope_resnet50}{65.0}{0.9} & \gradient{adaptiope_resnet50}{71.8}{0.9} \newtabsection \multirow{5}{*}{\gradientfull{officehome_vit}{86.4}{0.2}} & \gradient{adaptiope_vit}{49.2}{6.1} & \gradient{adaptiope_vit}{72.6}{9.1} & \gradient{adaptiope_vit}{84.6}{8.0} & \gradient{adaptiope_vit}{89.2}{6.8} \n
        
        & \textbf{2x32} \newtabsection & \gradient{adaptiope_resnet50}{46.6}{1.2} & \gradient{adaptiope_resnet50}{26.6}{0.3} & \gradient{adaptiope_resnet50}{62.2}{1.0} & \gradient{adaptiope_resnet50}{70.8}{0.9} \newtabsection & \gradient{adaptiope_vit}{49.2}{4.0} & \gradient{adaptiope_vit}{65.6}{1.8} & \gradient{adaptiope_vit}{86.3}{6.7} & \gradient{adaptiope_vit}{92.3}{0.7} \n
        
        & \textbf{4x16} \newtabsection & \gradient{adaptiope_resnet50}{32.3}{0.8} & \gradient{adaptiope_resnet50}{10.1}{0.6} & \gradient{adaptiope_resnet50}{59.5}{1.0} & \gradient{adaptiope_resnet50}{69.6}{0.9} \newtabsection & \gradient{adaptiope_vit}{38.6}{4.0} & \gradient{adaptiope_vit}{55.9}{3.8} & \gradient{adaptiope_vit}{88.3}{1.4} & \gradient{adaptiope_vit}{87.3}{6.9} \n
        
        & \textbf{8x8}  \newtabsection & \gradient{adaptiope_resnet50}{21.6}{1.2} & \gradient{adaptiope_resnet50}{10.2}{0.5} & \gradient{adaptiope_resnet50}{55.9}{1.6} & \gradient{adaptiope_resnet50}{67.9}{0.9} \newtabsection & \gradient{adaptiope_vit}{35.0}{4.1} & \gradient{adaptiope_vit}{51.5}{3.4} & \gradient{adaptiope_vit}{87.1}{1.4} & \gradient{adaptiope_vit}{76.4}{21.1} \n
        
        & \textbf{16x4} \newtabsection & \gradient{adaptiope_resnet50}{4.9}{1.2} & \gradient{adaptiope_resnet50}{18.9}{0.7} & \gradient{adaptiope_resnet50}{51.9}{2.3} & \gradient{adaptiope_resnet50}{66.0}{1.2} \newtabsection & \gradient{adaptiope_vit}{24.2}{9.1} & \gradient{adaptiope_vit}{42.8}{7.4} & \gradient{adaptiope_vit}{84.9}{6.9} & \gradient{adaptiope_vit}{82.4}{11.0} \n
        \midrule
        
        \multirow{5}{*}{\rotatebox[origin=b]{90}{\textbf{IMAGECLEF \hspace{0.2cm}}}} & \textbf{1x64} \newtabsection \multirow{5}{*}{\gradientfull{imageclef_resnet50}{81.1}{0.2}} & \gradient{imageclef_resnet50}{83.2}{0.2} & \gradient{imageclef_resnet50}{83.3}{0.1} & \gradient{imageclef_resnet50}{82.9}{0.2} & \gradient{imageclef_resnet50}{83.0}{0.3} \newtabsection \multirow{5}{*}{\gradientfull{imageclef_vit}{87.5}{0.0}} & \gradient{imageclef_vit}{87.8}{1.1} & \gradient{imageclef_vit}{88.4}{0.2} & \gradient{imageclef_vit}{87.8}{0.3} & \gradient{imageclef_vit}{87.7}{0.2} \n
        
        & \textbf{2x32} \newtabsection & \gradient{imageclef_resnet50}{82.8}{0.2} & \gradient{imageclef_resnet50}{82.9}{0.2} & \gradient{imageclef_resnet50}{82.7}{0.2} & \gradient{imageclef_resnet50}{82.8}{0.3} \newtabsection & \gradient{imageclef_vit}{88.2}{0.2} & \gradient{imageclef_vit}{88.4}{0.1} & \gradient{imageclef_vit}{87.6}{0.5} & \gradient{imageclef_vit}{86.2}{3.2} \n
        
        & \textbf{4x16} \newtabsection & \gradient{imageclef_resnet50}{82.5}{0.4} & \gradient{imageclef_resnet50}{82.7}{0.2} & \gradient{imageclef_resnet50}{82.7}{0.1} & \gradient{imageclef_resnet50}{82.7}{0.1} \newtabsection & \gradient{imageclef_vit}{87.1}{1.7} & \gradient{imageclef_vit}{88.0}{0.2} & \gradient{imageclef_vit}{87.6}{0.4} & \gradient{imageclef_vit}{87.4}{0.6} \n
        
        & \textbf{8x8}  \newtabsection & \gradient{imageclef_resnet50}{79.8}{1.6} & \gradient{imageclef_resnet50}{81.4}{0.2} & \gradient{imageclef_resnet50}{82.7}{0.2} & \gradient{imageclef_resnet50}{82.5}{0.1} \newtabsection & \gradient{imageclef_vit}{87.3}{0.8} & \gradient{imageclef_vit}{87.2}{0.2} & \gradient{imageclef_vit}{87.5}{0.3} & \gradient{imageclef_vit}{86.6}{1.9} \n
        
        & \textbf{16x4} \newtabsection & \gradient{imageclef_resnet50}{67.3}{2.0} & \gradient{imageclef_resnet50}{77.1}{0.7} & \gradient{imageclef_resnet50}{81.8}{0.1} & \gradient{imageclef_resnet50}{81.5}{0.2} \newtabsection & \gradient{imageclef_vit}{86.2}{0.9} & \gradient{imageclef_vit}{73.8}{2.1} & \gradient{imageclef_vit}{87.4}{0.2} & \gradient{imageclef_vit}{87.2}{0.4} \n
        \bottomrule
        
    \end{tabular}
\end{table*}

%% file: methods_average_results_table.tex
\begin{table}[h]
    \centering
    \captionsetup{width=0.9\linewidth}
    \caption{Average accuracy of different methods on Office31, Office-Home, Image-Clef, and Adaptiope. Each experiment (defined by a domain-pair of a dataset) is given equal importance in the average computation. The first row includes the baseline of Fine-Tuning without any adaptation. For each architecture best results are in \textbf{bold}, while other high and comparable results are \underline{underlined}.}
    \begin{tabular}{lcc}
        & \textbf{ResNet50} & \textbf{ViT-Large} \\
        \midrule
        \textbf{FT-ODG} &  65.52 & 85.95 \\
        \midrule
        \textbf{SHOT} & \underline{77.26} & \underline{88.40} \\
        \textbf{NRC}  & \underline{77.45} & 84.28 \\
        \textbf{AAD}  & 76.34 & 80.16 \\
        \textbf{PCSR} & \textbf{78.77} & \underline{88.12} \\
        \textbf{SCA}  & 71.74  & \textbf{88.42} \\
        \bottomrule
    \end{tabular}
    \label{tab:average_results}
\end{table}

%% file: summary_table.tex
\begin{table}[h]
\centering
\setlength{\tabcolsep}{2pt}
\caption{Results summary of the methods evaluation performed in Sec.~\ref{sec:methods} based on Office31, Office-Home, Adaptiope and Image-CLEF datasets and ResNet50 and ViT-Large Backbones.}

\begin{tabular}{lccccc}
\textbf{Methods} & \textbf{SCA} & \textbf{SHOT} & \textbf{NRC} & \textbf{AAD} & \textbf{PCSR} \\
\midrule
\textbf{Reprod.} & \textcolor{coolgreen}{\Large$\checkmark$} & \textcolor{coolgreen}{\Large$\checkmark$} & \textcolor{coolgreen}{\Large$\checkmark$} & \textcolor{coolgreen}{\Large$\checkmark$} & \textcolor{coolgreen}{\Large$\checkmark$} \\
\textbf{Parallelism} & \textcolor{coolgreen}{\Large$\checkmark$} & \textcolor{coolgreen}{\Large$\checkmark$} & \textcolor{coolred}{\Large \ding{55}} & \textcolor{coolred}{\Large \ding{55}} & \textcolor{coolgreen}{\Large$\checkmark$} \\
\textbf{Dataset Flex.} & \textcolor{coolgreen}{\Large$\checkmark$} & \textcolor{coolgreen}{\Large$\checkmark$} & \textcolor{coolgreen}{\Large$\checkmark$} & \textcolor{coolred}{\Large \ding{55}} & \textcolor{coolgreen}{\Large$\checkmark$} \\
\textbf{Backbone Flex.} & \textcolor{coolred}{\Large \ding{55}} & \textcolor{coolgreen}{\Large$\checkmark$} & \textcolor{coolred}{\Large \ding{55}} & \textcolor{coolred}{\Large \ding{55}} & \textcolor{coolgreen}{\Large$\checkmark$} \\
\textbf{Hparam-free} & \textcolor{coolgreen}{\Large$\checkmark$} & \textcolor{coolgreen}{\Large$\checkmark$} & \textcolor{coolred}{\Large \ding{55}} & \textcolor{coolred}{\Large \ding{55}} & \textcolor{coolgreen}{\Large$\checkmark$} \\
\bottomrule
\end{tabular}
\label{tab:summary}
\end{table}

%% file: experiments_pretraining.tex
    \begin{figure*}[t]
        \centering
        \includegraphics[width=0.98\linewidth]{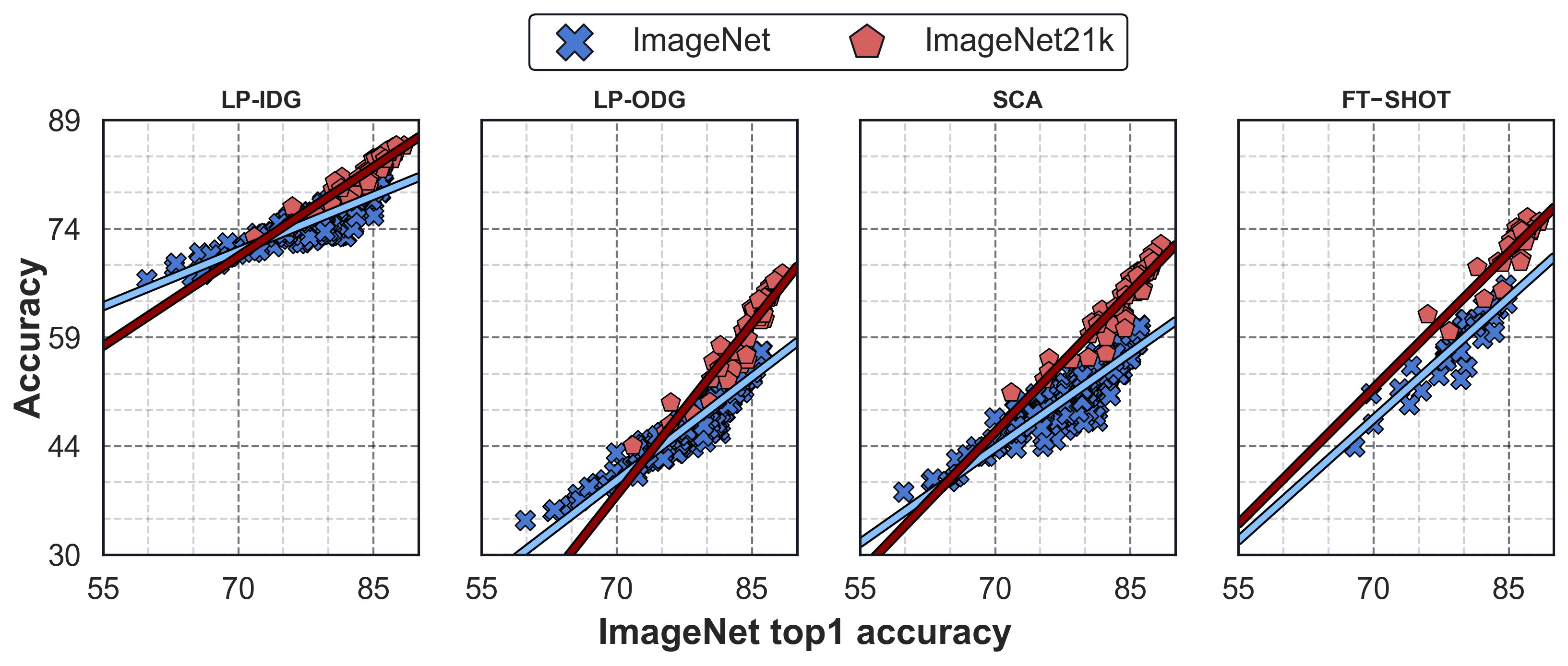}
        \caption{Left to right: LP-IDG accuracy (upper bound) averaged over 23 domains, LP-ODG (lower bound), SCA, and FT-SHOT accuracy (averaged over 74 domain pairs). Each marker indicates an architecture, with the x-axis denoting the ImageNet top1 accuracy. Markers color and shape signify the respective pre-training datasets. }
        \label{fig:four_plots_models}
\end{figure*}

This section aims to analyze the impact of strategic choices during pre-training on the final SF-UDA accuracy. 
To this purpose, we start with a comprehensive study considering over 500 backbones and the ImageNet and ImageNet21K datasets, to evaluate the impact of the backbone and dataset choices during pre-training (Sec.~\ref{subsec:statistics}). Note that we narrow down the analysis to a subset of 59 models for more computationally intensive evaluations, i.e., when fine-tuning (FT) is involved.
Finally, we study several self-supervised methods for pre-training and their effects on SF-UDA (Sec.~\ref{subsec:self-supervision}).

\subsection{Statistical analysis}
\label{subsec:statistics}
Fig.~\ref{fig:four_plots_models} reports the observed correlation between the ImageNet top-1 accuracy and the final accuracy on the task at hand of the two considered SF-UDA methods (SCA and FT-SHOT) and of their upper and lower bounds (LP-IDG and LP-ODG, respectively). Each different point represents a different backbone and we compare models trained on ImageNet (blue) with those trained on ImageNet21K (red).
The results show high correlation among the variables: models that achieve higher top-1 accuracy on ImageNet tend to generalize better across other tasks. 
While this connection had been previously noted for UDA~\citep{kornblith2019better}, our findings further stress its significance in the context of ODG and SF-UDA.
However, ImageNet top-1 accuracy, though indicative, is not the sole predictor of performance.
A distinct phenomenon can be observed with models pre-trained on the larger ImageNet21k dataset and subsequently fine-tuned on ImageNet. 
These models not only enhance their ImageNet performance but also exhibit additional improvements on SF-UDA tasks, not solely explained by their top-1 ImageNet accuracy gain.

To quantify the significance of the observed correlations, we present a statistical analysis identifying two predominant variables that can affect the final accuracy: the ImageNet top-1 accuracy and the choice of pre-training dataset.
If we ignore the effect of pre-training, the relationship for each backbone \( \mathcal{B} \) can be expressed as:

\begin{equation}
    \text{accuracy}(\mathcal{B}) = m \cdot \text{top1}(\mathcal{B}) + q + \epsilon,
\end{equation}

where \( m \) and \( q \) represent task-specific coefficients, and \( \epsilon \) is the residual variance not captured by the model. The choice of pre-training dataset is an important additional explanatory variable.
For simplicity, in our analysis we denote the pretraining dataset choice by a boolean variable 
\(\text{pretrain}(\mathcal{B})\), taking value \(0\) for ImageNet or \(1\) for ImageNet21k.
To account for the influence of the pre-training dataset on model performance, we employ a multi-linear model:

\begin{equation}
\begin{gathered}
    \text{accuracy}(\mathcal{B}) = [m + \Delta m \cdot \text{pretrain}(\mathcal{B})] \cdot \text{top1}(\mathcal{B}) + \\ + q + \Delta q \cdot \text{pretrain}(\mathcal{B}) + \epsilon,
\end{gathered}
\end{equation}

where \( \Delta m \) and \( \Delta q \) are the additional coefficients.
A comparison of both statistical models is illustrated in Tab.~\ref{table_R}, where it is possible to see a large increase of adjusted $\bar{R}^2$ values when cosidering multi-linear models. Finally, Tab.~\ref{table_R} shows that our findings are valid in all considered settings, with and without the fine-tuning.

\begin{table}[t]
\caption{Comparison of adjusted $\bar{R}^2$ values between linear (considering top-1 accuracy only) and multi-linear models (with both top-1 accuracy and pre-training). Results are shown for the upper bounds (LP-IDG and FT-IDG), the lower bounds (LP-ODG and FT-ODG) and for the considered SF-UDA methods, SCA and SHOT, with and without fine-tuning (FT).}
\label{table_R}

\setlength{\tabcolsep}{4.3pt}
\centering

\begin{tabular}{lccccc}

& \textbf{LP-IDG} & \textbf{LP-ODG} & \textbf{SCA} & \textbf{SHOT}\\
\midrule
\textbf{Lin} & 0.736 & 0.810 & 0.731 & 0.792\\
\textbf{M-Lin} & 0.851 & 0.935 & 0.902 & 0.890\\
\bottomrule
\end{tabular}

\vspace{0.3cm}
\centering
\setlength{\tabcolsep}{1.5pt}
\begin{tabular}{lccccc}
 & \textbf{FT-IDG} & \textbf{FT-ODG} & \textbf{FT-SCA} & \textbf{FT-SHOT}\\
\midrule
\textbf{Lin}  & 0.803 & 0.698 & 0.668 & 0.838\\
\textbf{M-Lin} & 0.878 & 0.822 & 0.792 & 0.932\\
\bottomrule
\end{tabular}

\end{table}

\textbf{Backbone size influence.} 
\cite{kolesnikov2020big} suggest that larger models, due to their extensive parameter set, would primarily benefit from pre-training on larger datasets like ImageNet21k, while smaller models can obtain no or negative benefit from very large-scale pre-trainings.
However, our investigation challenges this evidence in the context of SF-UDA. In Tab.~\ref{in_vs_21k}, we demonstrate that not only larger architectures, but even small and medium size backbones can significantly benefit from ImageNet21k pre-training.

\begin{table*}[t]
\caption{ImageNet vs ImageNet21k pre-training for different backbone sizes. The accuracy reported are the average of 74 domain shifts.
} 
\label{in_vs_21k}

\centering
\begin{tabular}{rcccccc}

& \multicolumn{2}{c}{\textbf{LP-ODG}} & \multicolumn{2}{c}{\textbf{SCA}} & \multicolumn{2}{c}{\textbf{FT-SHOT}} \\
\cmidrule{2-7}
\textbf{Backbone} (\textbf{params})& \textbf{IN} & \textbf{IN21k} & \textbf{IN} & \textbf{IN21k}& \textbf{IN} & \textbf{IN21k} \\
\midrule

VGG19 (143.7M)  & 45.2 & \textbf{47.6} & 49.4 & \textbf{53.1} & \textbf{56.0} & 55.7\\
ResNet50 (25.6M) & 47.2 & \textbf{51.3} & 49.8 & \textbf{58.2} & 55.9 & \textbf{62.0} \\
W-ResNet50 (68.9M) & 50.5 & \textbf{52.5} & 53.3 & \textbf{58.2} & 62.1 & \textbf{64.0} \\
DenseNet161 (28.7M) & 48.0 & \textbf{52.2} & 52.7 & \textbf{58.2} & 61.6 & \textbf{65.3} \\
ConvNeXt-Base (88.6M) & 54.4 & \textbf{65.2} & 58.4 & \textbf{68.5} & 65.1 & \textbf{72.7} \\
\bottomrule
\end{tabular}
\end{table*}

\subsection{Self-supervised pre-training} \label{subsec:ssl}
\label{subsec:self-supervision}

Recently, self-supervised pre-training has gained popularity due to its demonstrated effectiveness in many studies~\citep{gui2023survey}. In this section, we examine a set of well-known state-of-the-art self-supervised learning (SSL) pre-training methods within the SF-UDA context, comparing them to supervised pre-training.
Our focus is on the widely-employed ResNet50 and ViT-Base architectures, since they are frequently considered in SSL studies and their pre-trained weights are available online.

\noindent \textbf{SSL with ResNet50.} For the analysis with ResNet50, we evaluate the DINO~\citep{dino}, MOCO v1~\citep{mocov1}, and MOCO v2~\citep{mocov2} approaches. As shown in Tab.~\ref{tab:ssl_resnet50}, when considering FT-ODG and FT-SCA, both MOCO v2 and supervised pre-training on ImageNet (1k) exhibit similar results, while other methods perform worse. 
However, when applying SHOT after the fine-tuning phase (FT-SHOT), supervised pre-training clearly outperforms all SSL pre-training methods, including MOCO v2. A detailed analysis of why FT-SCA does not perform as well as FT-SHOT in this context is presented in Sec.~\ref{sec:finetuning}. Thus, for ResNet50, supervised pre-training is more effective than the considered SSL methods.

\begin{table*}[t]

\caption{Target accuracy (\%) using different SSL initialization weights of ResNet50 after performing the following experiments: fine-tuning on the source domain (FT-ODG), FT-SCA and FT-SHOT.}

\label{tab:ssl_resnet50}
\centering

\setlength{\tabcolsep}{3pt}
\begin{tabular}{llccccccc}

\textbf{Exp.} & \textbf{Pre-train} & M-Office31 &  Visda &  O-Home &  Adaptiope &  I-CLEF &  D.Net &  \textbf{Avg}\\
\midrule
\multirow{4}{*}{FT-ODG} & DINO     &  42.7 &   45.6 &        45.7 &       30.9 &     71.3 &       19.7 & 42.7\\
                           & MOCO v1  &  38.5 &   44.6 &        41.7 &       28.6 &     66.2 &       20.1 & 40.0\\
                           & MOCO v2  &  54.7 &   \textbf{54.2} &        55.1 &       39.5 &     \textbf{76.9} &       \textbf{25.5} & \textbf{51.0}\\
                           & Supervised (IN1k)  &  \textbf{56.4} &   49.6 &        \textbf{55.5} &       \textbf{43.3} &     75.9 &       21.6 & 50.4\\
\midrule
\multirow{4}{*}{FT-SCA}    & DINO     &  50.8 &   54.5 &        49.8 &       35.9 &     75.0 &       21.9 & 48.0\\
                           & MOCO v1  &  49.5 &   52.3 &        43.9 &       34.3 &     70.4 &       23.9 & 45.7\\
                           & MOCO v2  &  \textbf{63.4} &   \textbf{56.9} &        55.8 &       45.3 &     \textbf{79.5} &       \textbf{28.0} & 54.8\\
                           & Supervised (IN1k) &  62.7 &   55.5 &        \textbf{59.4} &       \textbf{49.6} &     79.3 &       23.6 & \textbf{55.0}\\
\midrule
\multirow{4}{*}{FT-SHOT}   & DINO     &  71.8 &   67.3 &        58.0 &       55.4 &     77.5 &       24.1 & 59.0\\
                           & MOCO v1  &  56.6 &   58.3 &        52.9 &       37.5 &     71.7 &       19.2 & 49.4\\
                           & MOCO v2  &  73.2 &   65.7 &        66.9 &       51.9 &     81.0 &       26.2 & 60.8\\
                           & Supervised (IN1k)  &  \textbf{80.6} &   \textbf{71.6} &        \textbf{68.9} &       \textbf{66.6} &     \textbf{81.6} &       \textbf{27.3} & \textbf{66.1}\\
                           
\bottomrule
\end{tabular}
\end{table*}

\begin{table*}[t]
\centering
\caption{
Target accuracy (\%) for VIT-Base we report results for the lower bounds (LP-ODG and FT-ODG) and for the considered SF-UDA methods (SCA and SHOT) with and without fine-tuning (FT). DINO v2 and MAE need distinct FT strategies (indicated with \textsuperscript{\textdagger}) and their backbones were frozen during SHOT adaptation (indicated with an \textsuperscript{*}).  Best results for each dataset and setting are highlighted in \textbf{bold}, with the top overall results \underline{underlined}.
}

\label{tab:ssl_vitbase}

\scriptsize
\begin{tabular}{llcccc}

\textbf{Experiment} & \textbf{Pre-train} & Office31 & Modern Office31 &  Visda &  Office-Home \\
\midrule
\multirow{4}{*}{LP-ODG} & MAE & 44.2 & 30.0 & 45.6 & 37.8 \\
                        & DINO & 81.5 & 71.6 & 65.7 & 59.8 \\
                        & DINO v2 & \textbf{89.4} & \textbf{89.8} & \textbf{72.6} & \textbf{79.3} \\
                        & Supervised (IN21k) & \textbf{89.4} & 85.1 & 69.4 & 78.2 \\
\midrule

\multirow{4}{*}{FT-ODG} & MAE\textsuperscript{\textdagger}  & 52.9 & 39.0 & 53.4 & 51.0 \\
                        & DINO & 80.7 & 72.4 & 65.4 & 63.6 \\
                        & DINO v2\textsuperscript{\textdagger}  & \textbf{89.8} & \textbf{87.7} & 58.5 & 76.8 \\
                        & Supervised (IN21k) & 89.7 & 86.9 & \textbf{77.9} & \textbf{79.9} \\

\midrule

\multirow{4}{*}{SCA}   & MAE & 16.4 & 15.6 & 17.6 & 5.4 \\    
                       & DINO &  86.6 & 84.4 & 74.2 & 61.1 \\
                       & DINO v2 & \textbf{94.1} & \underline{\textbf{96.5}} & \textbf{81.7} & \textbf{83.0} \\
                       & Supervised (IN21k) & 92.7 & 91.8 & 69.1 & 78.5 \\
                    
\midrule

\multirow{4}{*}{FT-SCA}    & MAE\textsuperscript{\textdagger}  & 34.9 & 28.7 & 49.8 & 32.0 \\    
                           & DINO &  85.2 & 81.1 & 73.6 & 67.9 \\
                           & DINO v2\textsuperscript{\textdagger}  &  \textbf{93.8} & \textbf{95.0} & 62.8 & 79.8 \\
                           & Supervised (IN21k) & 92.7 & 91.9 & \textbf{82.6} & \textbf{82.5} \\

\midrule

\multirow{4}{*}{SHOT}      & MAE* & 8.4 & 12.4 & 62.2 & 18.1 \\   
                           & DINO & 89.4 & 86.5 & 79.2 & 68.3 \\
                           & DINO v2* &  94.0 & 96.8 & 80.6 & 84.7 \\
                           & Supervised (IN21k) & \underline{\textbf{95.3}} & \textbf{93.9} & \underline{\textbf{88.5}} & \underline{\textbf{86.5}} \\

\midrule

\multirow{4}{*}{FT-SHOT}   & MAE\textsuperscript{\textdagger *} & 38.0 & 39.0 & 67.1 & 55.7 \\   
                           & DINO & 87.5 & 75.8 & 56.9 & 74.2 \\
                           & DINO v2\textsuperscript{\textdagger *} &  93.3 & 95.5 & 68.4 & 81.2 \\
                           & Supervised (IN21k) & \textbf{93.7} & \textbf{93.9} &  \textbf{87.5} & \textbf{84.4} \\

\bottomrule
\end{tabular}
\end{table*}

\noindent \textbf{SSL with ViT.} For the ViT-Base backbone, we consider Masked Autoencoder (MAE)~\citep{he2022masked}, DINO~\citep{dino}, and DINO v2~\citep{oquab2023dinov2} as SSL pre-training strategies.
The results are presented in Tab.~\ref{tab:ssl_vitbase}. 
Despite recognizing the effectiveness of MAE, our results show that, without a subsequent supervised ImageNet fine-tuning, it does not provide a strong foundation for SF-UDA and ODG tasks.
Nevertheless, DINO  shows stronger performance than MAE on all the examined tasks, although it is still outperformed by supervised pre-training.

\noindent Finally, for the DINO v2 evaluation
we faced two main challenges.
The first is related to its fine-tuning on the source domain, caused by gradients of high magnitude in the initial training phase that hardly alter the weights of the model reducing the inherent benefits of DINO v2 initialization.
To address this issue, we propose to adopt a two-phase fine-tuning strategy: (i)~we freeze the backbone, training only the newly introduced bottleneck and classifier on the source domain, and (ii)~we continue the whole network fine-tuning using gradient clipping. 
The second challenge is related to adaptation to the target domain with SHOT.
SHOT adaptation fails with DINO~v2 when the entire network is optimized with the unsupervised objective.
In our pipeline, to solve this problem we freeze the DINO~v2 weights and only optimize the bottleneck weights with  SHOT. 
Applying FT-SHOT on DINO v2 without our approach leads to bad performance on all datasets. For instance, on Office31, the accuracy is $5.6 \%$ without our method, while it is $93.3\%$ when we apply it. Analogous patterns are evident on Modern Office31 (from $7.0\%$ to $95.5\%$), VisDA (from $14.9\%$ to $68.4\%$), and Office-Home (from $3.5\%$ to $81.4\%$). 
In general, if our approach is used on DINO v2, it obtains similar performance to supervised pre-training, even surpassing it in some settings.

\noindent Note that we apply the same strategies presented for DINO v2 to MAE.
This yields marginal improvements, but even with these enhancements the performance remains sub-optimal, rendering MAE less competitive than the other methods.

%% file: experiments_finetuning.tex
\begin{figure*}[h!]
    \centering
    \includegraphics[width=0.94\linewidth]{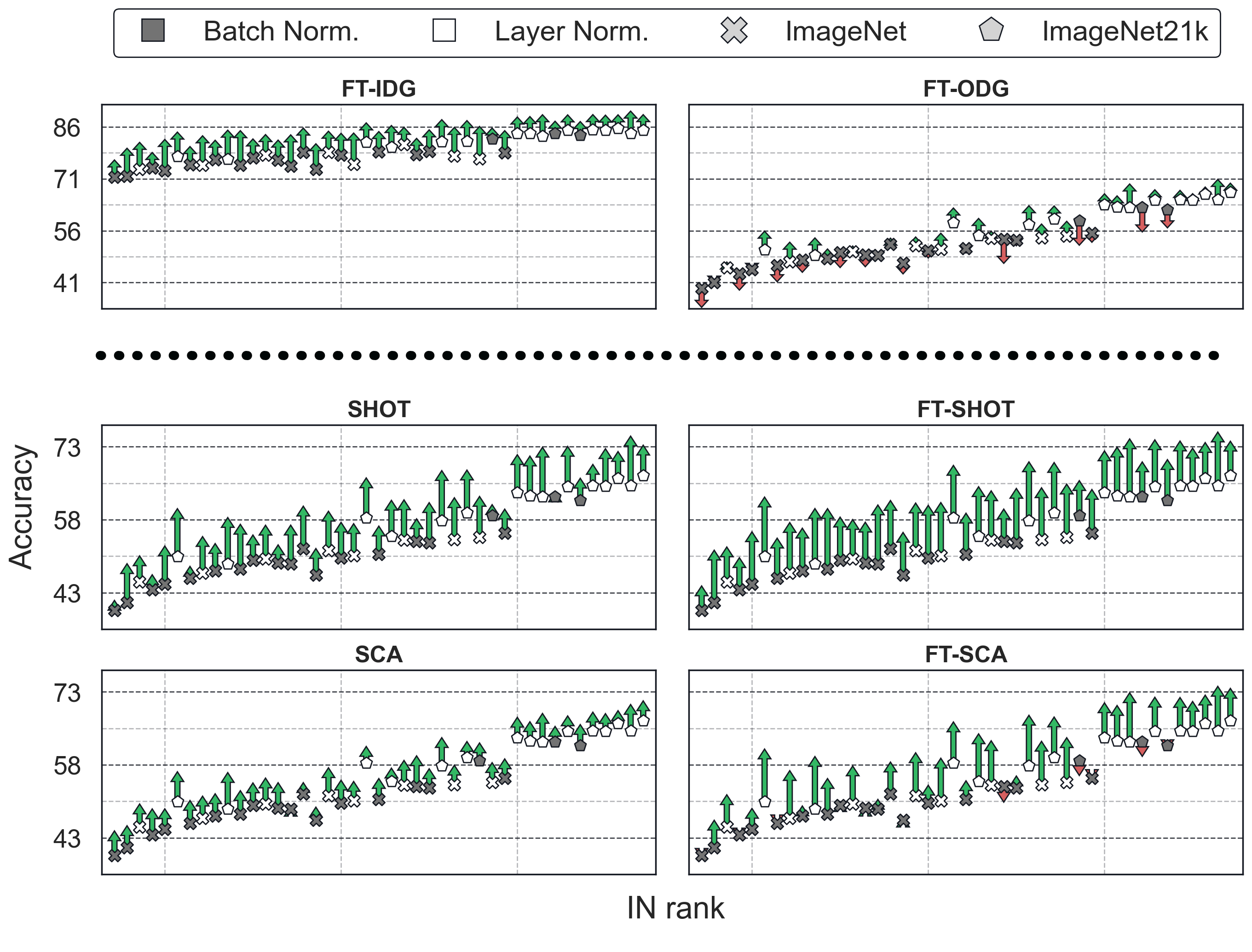}
\caption{
Impact on final accuracy of source fine-tuning. Each marker is a backbone. Variations on the final accuracy is reported as arrows. In the top row, we report the accuracy difference between LP-IDG and FT-IDG (left) and LP-ODG and FT-ODG (right), being our upper and lower bounds, respectively. All other plots represent accuracy difference between LP-ODG with the considered SF-UDA method (i.e., SHOT, FT-SHOT, SCA and FT-SCA). Models with BN and LN are represented with dark and white markers, respectively.}

\label{fig:ft_vs_noft}
\end{figure*}

In SF-UDA, the backbone is typically fine-tuned on the source domain during the first transfer.
Given the importance of this step, we analyze its impact on the overall SF-UDA performance.
To this aim, we study the accuracy variation happening when fine-tuning (FT) is applied with respect to cases without it. 
We report the results in Fig.~\ref{fig:ft_vs_noft}, where each marker represents a backbone, sorted on the X-axis by its top-1 accuracy on ImageNet (best models are on the right). The Y-value of each marker indicates the baseline LP (IDG or ODG) accuracy of that backbone, while the arrows represent the accuracy variation occurring for the model when a method is applied.
In the top-left plot, we report the accuracy comparison
between LP-IDG (markers) and FT-IDG (arrows), while in the top-right plot we report the comparison between LP-ODG (markers) and FT-ODG (arrows), being our
upper and lower bounds, respectively. The plots in the other rows represent the accuracy comparison between the lower bound (LP-ODG) and the considered SF-UDA methods (i.e., SHOT, FT-SHOT, SCA, and FT-SCA).
Thus, the impact of FT on SF-UDA methods can be evaluated by comparing the two plots in the second and third rows.
As shown in Fig.~\ref{fig:ft_vs_noft} and reported in Tab.~\ref{table:delta_acc_and_failures} (left), SHOT combined with fine-tuning (FT-SHOT) consistently enhances performance across different backbones. 
Instead, for SCA, the benefit of fine-tuning (FT-SCA) is particularly evident for LN models with an increase of $8.14\%$, while the improvement for BN models is modest ($2.16\%$).
A similar pattern emerges for the lower bound FT-ODG, where the accuracy increases by $0.46\%$ across all architectures for all the experiments.
However, backbones with Batch Normalization (BN) and with Layer Normalization (LN) display different behaviors.
For the first ones, the final accuracy declines by $2.21\%$, while for the second ones, it improves by $3.62\%$.
Finally, the upper bound (FT-IDG) does not show the same pattern. This is due to the fact that the FT impact in the same domain is beneficial for both BN and LN models, as already well-established~\citep{kornblith2019better}.

\begin{table*}
    \centering

    \setlength\tabcolsep{4pt}

    \caption{Average accuracy variation and failure rates with respect to LP-ODG (lower bound). These are averaged over 74 domain shifts. Each column represents a normalization layer type. The parenthesis shows the number of considered backbones.} \label{table:delta_acc_and_failures}
        
    \begin{tabular}{lcccccc}

    & \multicolumn{3}{c}{$\Delta$ \textbf{Accuracy}} & \multicolumn{3}{c}{\textbf{Failure Rate}} \\
    \cmidrule(lr){2-4} \cmidrule(lr){5-7}
    \textbf{Tasks} & \textbf{All} & \textbf{BN (32)} & \textbf{LN (27)} & \textbf{All} & \textbf{BN (32)} & \textbf{LN (27)} \\
    \midrule
    \textbf{FT-ODG}    & $0.46$ {\footnotesize $\pm 3.71$} & $-2.21$ {\footnotesize $\pm 2.75$} & $3.62$ {\footnotesize $\pm 1.57$} & $40.82$ {\footnotesize $\pm 16.14$} & $51.52$ {\footnotesize $\pm 11.42$} & $28.13$ {\footnotesize $\pm 10.85$} \\
    \textbf{SCA}       & $4.21$ {\footnotesize $\pm 1.39$} & $4.08$ {\footnotesize $\pm 1.55$} & $4.36$ {\footnotesize $\pm 1.19$} & $21.05$ {\footnotesize $\pm 11.92$} & $27.24$ {\footnotesize $\pm 11.64$} & $13.71$ {\footnotesize $\pm 7.25$} \\
    \textbf{SHOT}      & $6.97$ {\footnotesize $\pm 2.24$} & $6.21$ {\footnotesize $\pm 2.39$} & $7.88$ {\footnotesize $\pm 1.68$} & $11.18$ {\footnotesize $\pm 8.75$} & $14.82$ {\footnotesize $\pm 9.32$} & $6.86$ {\footnotesize $\pm 5.61$} \\
    \textbf{FT-SCA}    & $4.90$ {\footnotesize $\pm 3.86$} & $2.16$ {\footnotesize $\pm 2.94$} & $8.14$ {\footnotesize$\pm 1.66$} & $19.08$ {\footnotesize $\pm 14.57$} & $29.77$ {\footnotesize $\pm 11.05$} & $6.41$ {\footnotesize $\pm 4.52$} \\
    \textbf{FT-SHOT}   & $9.66$ {\footnotesize $\pm 1.83$} & $9.68$ {\footnotesize $\pm 1.96$} & $9.63$ {\footnotesize $\pm 1.69$} & $4.47$ {\footnotesize $\pm 3.61$} & $5.83$ {\footnotesize $\pm 3.77$} & $2.85$ {\footnotesize $\pm 2.69$} \\
    \bottomrule
    \end{tabular}

\end{table*}

\subsection{Layer or Batch Normalization?}

\begin{figure*}[h!]
    \centering
    \includegraphics[width=0.94\linewidth]{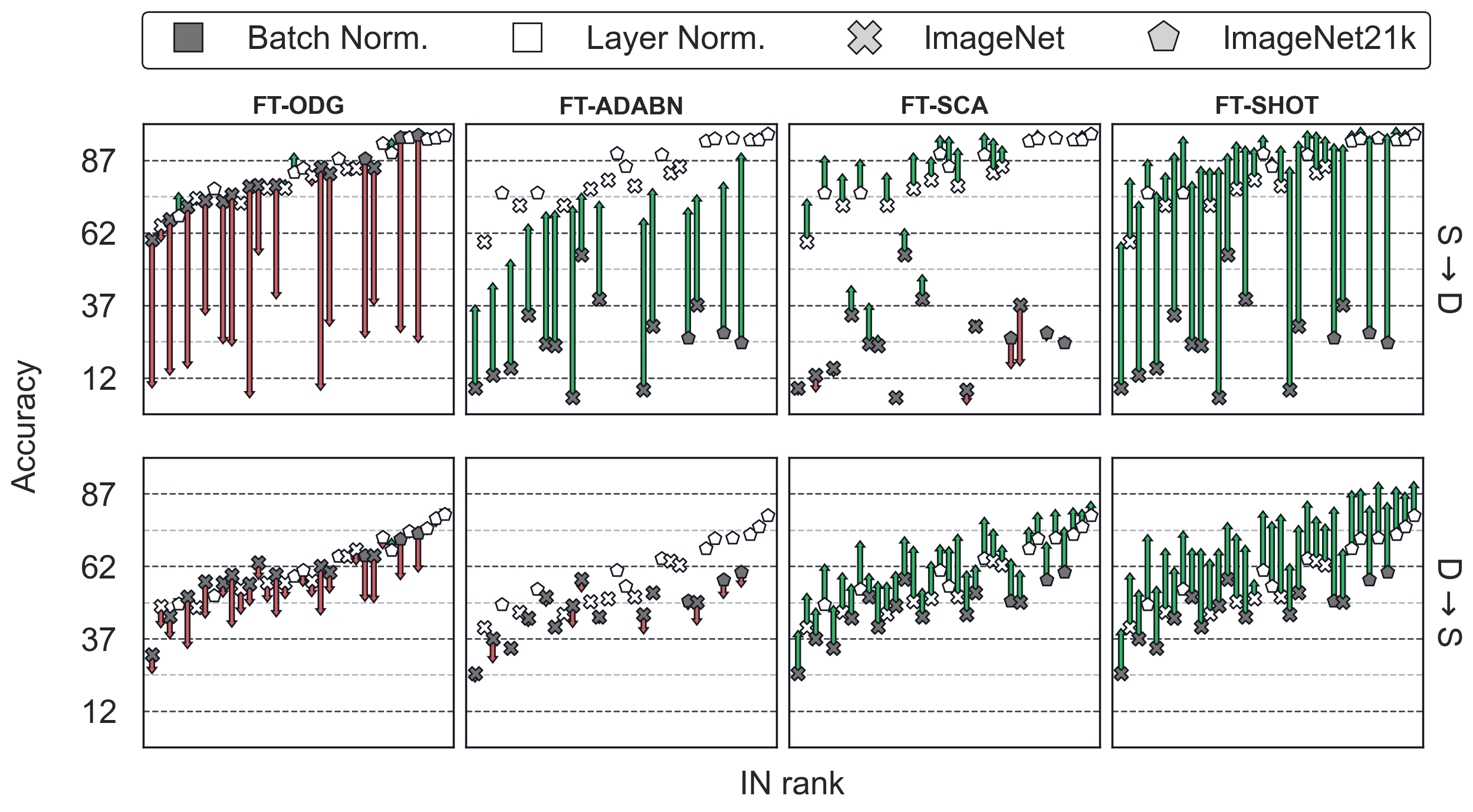}
    \caption{
    Example of fine-tuning impact for the Modern Office31 dataset (Synthetic $\to$ DSLR in the first row and DSLR $\to$ Synthetic in the second row). Fine-tuning effect on the final accuracy is reported as arrows. In the first column, we report the accuracy variation between LP-ODG and FT-ODG (lower bound). Then, we report the accuracy difference with FT-ODG for the case when ADABN is applied (second column) and for FT-SCA and FT-SHOT (third and fourth columns).}
    \label{fig:plot_example_failures}
\end{figure*}

In this section, we analyze the failure rate for each backbone after the fine-tuning on the source domain. 
We define a failure to be a performance degradation with respect to the LP-ODG lower bound when a method is applied.

Starting from the previously presented findings, in this section we compare backbones with LN and with BN.
We report our results in Tab.~\ref{table:delta_acc_and_failures} (right side). It is evident that models with BN consistently show higher failure rates compared to those with LN. For instance, for FT-SCA, models with BN have a 29.77\% failure rate, compared to only 6.41\% for models with LN.

Without source fine-tuning, SHOT, which adjusts BN statistics and backbone weights during adaptation, results in a 14.82\% BN failure rate.
In contrast, SCA achieves a failure rate of 27.24\% for BN models.
However, LN models failure rates significantly decrease for both SHOT and SCA, attaining 6.86\% and 13.71\%, respectively.
This large improvement underscores BN's instability in the presence of domain shifts, in contrast with LN models' robustness.

However, the literature presents techniques to mitigate BN limitations, such as ADABN~\citep{li2016revisiting}. Our findings reveal that, while these techniques can, in some cases, recover part of the lost accuracy, they remain inconsistent.
We report our findings in Fig.~\ref{fig:plot_example_failures}
for the \textit{DSLR} and \textit{Synthetic} domain pair in the Modern Office 31 dataset, taken as a representative example. 
ADABN occasionally falls short of a complete recovery. 
Additionally, as depicted in the second row, in some situations (i.e., specific source-target pairs), ADABN amplifies the degradation, raising questions about the scenarios in which such techniques might be beneficial.

Therefore, while various techniques can sometimes enhance BN models performance, their unpredictability across different domains often makes them less appealing for SF-UDA. 
LN models, on the contrary, demonstrate greater stability and reduced failure rates, rendering them a more reliable choice for SF-UDA tasks.

%% file: Discussions.tex
\begin{figure*}[h!]
        \centering
        \includegraphics[width=0.94\linewidth]{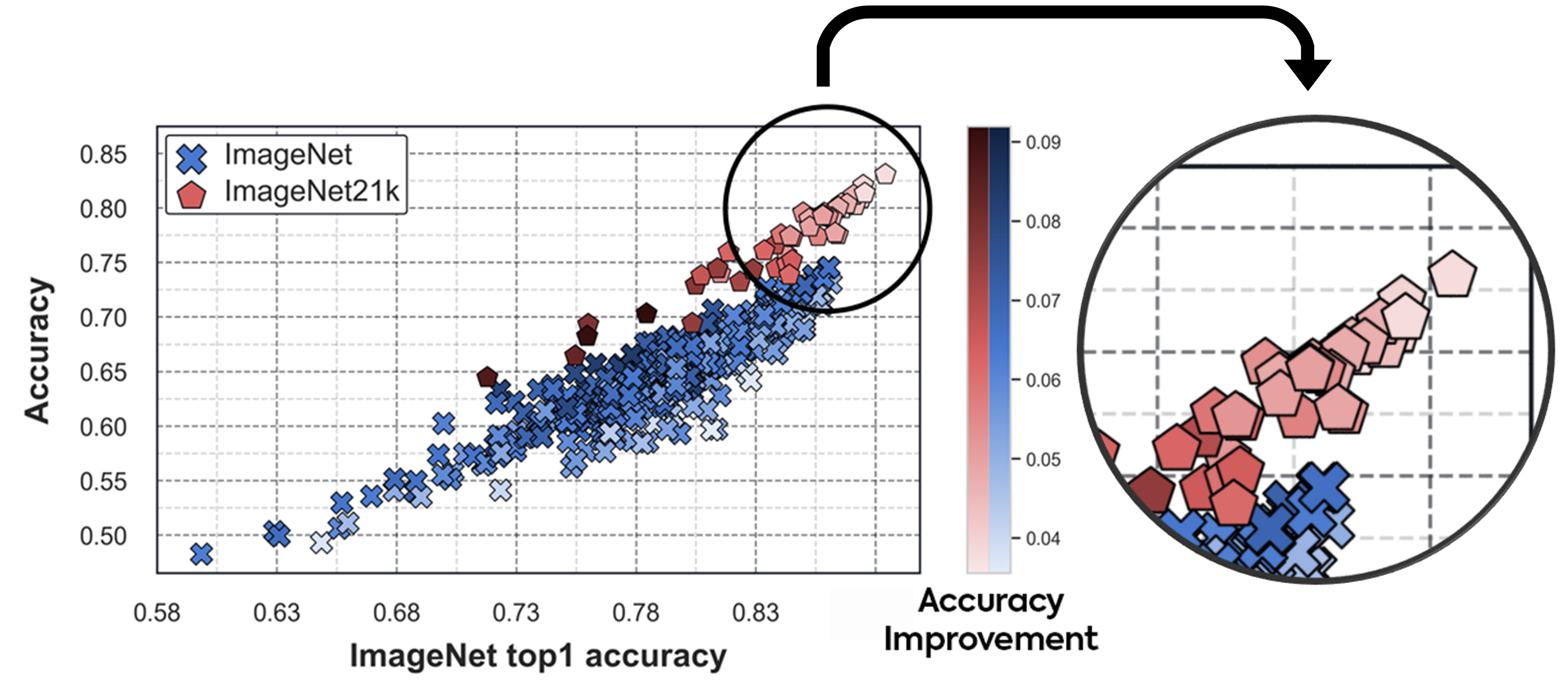}
        \caption{ We report the relation between the ImageNet top-1 accuracy and target accuracy of SCA across 74 domain pairs, for over 500 different backbones. Different colors of the markers represent different pre-training dataset, while the shade intensity signifies the SCA accuracy improvement with respect to the lower bound LP-ODG.}
        \label{fig:future_sfuda}
\end{figure*}

Our work proposes an open-source benchmark framework that enables the execution of a systematic large-scale experimental analysis of SF-UDA. 
We believe that this framework might put the basis for future work, being a useful tool for structured SF-UDA evaluation. 
Moreover, we present a detailed study of current SF-UDA state-of-the-art methods, highlighting relevant findings and useful insights. Firstly, in Sec.~\ref{sec:methods}, we point to limitations in prevalent and common benchmark standards.
We study the reproducibility, possibility to execute distributed training, dependence on datasets and backbones and hyperparameters sensitivity of SF-UDA methods. 
In this perspective, we believe that the method design and hyperparameter selection should not be driven by test accuracy after trial and error experiments. 
Otherwise, SF-UDA methods effectiveness and generality might be potentially overestimated.

Then, in Sec.~\ref{sec:pretraining}, we analyze the impact of the pre-training strategy and backbone choice for SF-UDA.
Architectures that perform well on ImageNet also present improved SF-UDA accuracy.
Pre-training on larger datasets, such as ImageNet21k, amplifies this outcome.
Although certain Self-Supervised Learning methods might not match the efficacy achieved by supervised training in general, SSL techniques such as DINO~v2 achieve competitive results (Sec.~\ref{subsec:ssl}).
However, an important question is still open: as architectures get larger and with improved  performance on ImageNet, is there a diminishing return on the advantages conferred by SF-UDA?
Preliminary observations, captured in Fig.~\ref{fig:future_sfuda}, suggest that this might be the case, since the best ImageNet21k models often benefit less from SF-UDA. 
Finally, in Sec.~\ref{sec:finetuning}, we provide an analysis on the effects of fine-tuning the backbone weights on the source domain during the first transfer.
While it is often beneficial, there exist scenarios, particularly involving backbones with Batch Normalization layers, where it can lead to hard failures.

\noindent \textbf{Limitations.} Although extensive, our study does not encompass the entire spectrum of SF-UDA methodologies.
We select a subset of them, emphasizing those that we found most representative, reproducible, and applicable to different datasets and backbones.
Some relevant aspects such as the evaluation of normalization-free networks~\citep{brock2021high} or with alternative normalization layers, the pre-training on different, larger, datasets and unsupervised hyperparameter selection for SF-UDA remain open for future research.

%% file: Conclusions.tex
In our investigation, we address many critical aspects of the SF-UDA landscape, offering an analysis on current practices and directions for forthcoming research.
In particular, we  highlight the strengths and weaknesses   of several representative SF-UDA methods, the role of the first transfer, and the importance of architectural choices and pre-training strategies, which undeniably show a significant impact on downstream SF-UDA performance.
In essence, this work represents a pivotal reference providing extensive and in-depth findings and guidance for researchers in SF-UDA and practitioners aiming to apply SF-UDA methods to real-world problems.

%% file: data_availability.tex
The supplementary material accompanying this paper contains the results of all experiments conducted to achieve the findings reported herein. It also includes the necessary code to replicate these experiments. The datasets employed in this study are exclusively available from official sources and can be freely accessed online. 

%% file: acknowledgements.tex
This study was carried out within the FAIR - Future Artificial Intelligence Research and received funding from the European Union Next-GenerationEU (PIANO NAZIONALE DI RIPRESA E RESILIENZA (PNRR) – MISSIONE 4 COMPONENTE 2, INVESTIMENTO 1.3 – D.D. 1555 11/10/2022, PE00000013). 
The paper was supported by the Italian National Institute for Insurance against Accidents at Work (INAIL) ergoCub Project.
L. R. acknowledges the financial support of the European Research Council (grant SLING 819789), the European Commission (ELIAS 101120237), the US Air Force Office of Scientific Research (FA8655-22-1-7034), the Ministry of Education, University and Research (grant ML4IP R205T7J2KP) and the Center for Brains, Minds and Machines (CBMM), funded by NSF STC award CCF-1231216.
This manuscript reflects only the authors’ views and opinions, neither the European Union nor the European Commission nor other organizations can be considered responsible for them.